\documentclass[nohyperref]{article}

\usepackage{microtype}
\usepackage{graphicx}
\usepackage{subfigure}
\usepackage{booktabs} 

\usepackage{hyperref}

\usepackage[accepted]{icml2022}

\usepackage{amsmath}
\usepackage{amssymb}
\usepackage{mathtools}
\usepackage{amsthm}
\newcommand*{\defeq}{\stackrel{\text{def}}{=}}
\usepackage[capitalize,noabbrev]{cleveref}

\theoremstyle{plain}

\theoremstyle{definition}

\theoremstyle{remark}

\newcommand{\norm}[1]{\left\lVert#1\right\rVert}
\usepackage[textsize=tiny]{todonotes}

\icmltitlerunning{Hierarchical Strategies for Cooperative Multi-Agent Reinforcement Learning}

\begin{document}

\twocolumn[
\icmltitle{Hierarchical Strategies for Cooperative Multi-Agent Reinforcement Learning}



\icmlsetsymbol{equal}{*}

\begin{icmlauthorlist}
\icmlauthor{Majd Ibrahim}{equal,hiast}
\icmlauthor{Ammar Fayad}{equal,mit}

\end{icmlauthorlist}
\icmlaffiliation{hiast}{HIAST}
\icmlaffiliation{mit}{MIT}

\icmlcorrespondingauthor{Ammar Fayad}{afayad@mit.edu}
\icmlcorrespondingauthor{Majd Ibrahim}{majd.ibrahim@hiast.edu.sy}

\icmlkeywords{Coordination, Cooperation, Reinforcement Learning, ICML, Intrinsic Motivation}

\vskip 0.3in
]



\printAffiliationsAndNotice{\icmlEqualContribution} 

\begin{abstract}
Adequate strategizing of agents’ behaviors is essential to solving cooperative MARL problems. One intuitively beneficial yet uncommon method in this domain is predicting agents’ future behaviors and planning accordingly. Leveraging this point, we propose a two-level hierarchical architecture that combines a novel information-theoretic objective with a trajectory prediction model to learn a “strategy”. To this end, we introduce a latent policy that learns two types of latent strategies: individual $z_A$, and relational $z_R$ using a modified Graph Attention Network module to extract interaction features. We encourage each agent to behave according to the strategy by conditioning its local $Q$-functions on $z_A$, and we further equip agents with a shared $Q$-function that conditions on $z_R$ . Additionally, we introduce two regularizers to allow predicted trajectories to be accurate and rewarding. Empirical results on Google Research Football (GRF) and StarCraft (SC) II micromanagement tasks show that our method establishes a new state of the art being, to the best of our knowledge, the first MARL algorithm to solve all super hard SC II scenarios as well as the GRF full game with a win rate higher than $95\%$, thus outperforming all existing methods. Videos and brief overview of the methods and results are available at: \href{https://sites.google.com/view/hier-strats-marl/home}{https://sites.google.com/view/hier-strats-marl}
\end{abstract}

\section{Introduction}
Considering that many real-life applications can be modeled as multi-agent systems (coordination of robot swarms \citep{huttenrauch2017guided}, intelligent warehouse systems \citep{nowe2012game}, autonomous navigation \citep{cao2012overview}, traffic network optimization \citep{stolle2002learning}), this research area has been receiving increasing attention in recent years. Cooperative multi-agent reinforcement learning (MARL) stands as a key tool for addressing challenges in  such systems, allowing agents to coordinate their behaviors to reach better performances. In recent years, many cooperative MARL methods have investigated   various challenges that hinder the convergence of agents' policies, amongst which one can mention: the scalability challenge (the joint action-observation space growing exponentially with the number of agents), partial observability, and communication constraints.

Additional challenges arise in environments with extreme stochasticity and sparse rewards which makes the learning process slow and particularly hard to establish efficient coordination in multi-agent tasks. 
With that in mind, an RL research direction emerged aiming  to develop effective hierarchy induction methods that can acquire lower-level policies that are built upon by higher-level policy that operates at a coarser level of temporal abstraction, where the latter is meant to learn intelligent behaviors and can reason at a higher level of abstraction to solve more complex tasks \citep{co2018self}. 
Moreover, the lower-level policies could act as learning agents in the environment, equipped with standard learning approaches. 
In the multi-agent settings, multiple works adopted a hierarchical structure that tends to address an inherited limitation in the base learning algorithm \citep{mahajan2019maven} or to learn high-level skills or roles \citep{wang2021rode,wang2020roma,pmlr-v139-liu21m}. These methods showed improvements with better performances. However, as the difficulty level of the environment increases, 
these methods fail to establish sophisticated coordination with consistent exploration, thus producing poor performances (as demonstrated in our GRF experiments). A plausible explanation to this phenomenon is that these learning algorithms face difficulties establishing cooperation at the level of noisy low-level actions due to extreme stochasticity and perform poor planning in the sense of long credit assignment with the existence of highly delayed rewards
which leads to drastic failures in the learning process \citep{yang2020hierarchical}.
In this paper, we attempt to push forward the progress on solving these dilemmas by introducing HISMA, a novel hierarchical multi-agent RL algorithm that acquires a set of strategies by predicting the agents' long-term futures.

First, we propose to construct a continuous latent space of strategies instead of a discrete set of skills, goals, or behaviors by encoding the agents' local trajectories through a novel information-theoretic objective. Second, we introduce a probabilistic latent variable model (which we refer to as the \textit{latent policy}), and train it to produce strategies that are consistent with agents' information and can predict their outcome. Furthermore, we incentivize our latent policy to exercise high-level reasoning not only to reflect agents' predicted futures, but also to generate latents that yield optimal performances. We define agents' policies as latent-conditioned decoders since they primarily execute the strategy devised by the latent policy, and show that we can train them using any fully-cooperative MARL algorithm. 

To that end, several challenges arise that might obstruct learning: the unavailability of sufficient data  about agents' trajectories thus inaccurate predictive models; and the fact that agents' interactions significantly affect their behaviors. We addressed the former by intrinsically motivating the latent-conditioned policies to explore diverse data via a fully unsupervised procedure. The latter challenge was addressed by learning a two-fold latent strategy for each agent: one that encodes an agent's self-intentions and behavior, while the other represents the agent's social behavior. The latter is referred to as a \textit{relational} strategy and is designed to learn interaction features via augmenting a modified graph attention network to leverage coherence in agents' behaviors.

In summary, we make the following key contributions: \textbf{1)} we introduce a novel hierarchical multi-agent reinforcement learning algorithm that acquires a continuous latent space of strategies, together with a predictive model that can reflect the outcomes of those strategies; 
\textbf{2)} we propose a novel information-theoretic objective to maintain agents' awareness of their strategies as well as consistency between learned strategies and agents' behaviors (including predicted futures); 
\textbf{3)}  we improve exploration and diverse data collection by introducing a novel \textit{surprise} measure as an intrinsic motivation to the learning agents;
\textbf{4)} we empirically show the cutting-edge performance of our method by evaluating it on the most challenging and complex environments (e.g. SC II and GRF), and demonstrate that HISMA pushes forward the state of the art in MARL since it outperforms all powerful methods in the field, sometimes by a wide margin. 


\section{Background}
We formulate a fully cooperative multi-agent task as a Decentralized Partially Observable MDP (Dec-POMDP) \citep{10.5555/2967142}, which is defined as a tuple $\mathcal{M}=\langle \mathcal{N}, \mathcal{S},\mathcal{A}, \mathcal{O}, P, \Omega, n, \gamma \rangle$, where $\mathcal{N}$ is a finite set of $n$ agents, $s\in\mathcal{S}$ is the global state of the environment. At each time step $t$, every agent $i \in \mathcal{N}$ chooses an action
$a^i_t \in \mathcal{A}$ on a global state $s_t$, which forms a joint action $\mathbf{a}_t= (a^1_t,a^2_t,\dots,a^n_t)\in \mathcal{A}^n $.
It results in a joint reward $r(s_t,\mathbf{a}_t)$ and a transition to the next global state $s_{t+1} \sim P(.|s_t,\mathbf{a}_t)$. $\gamma \in [0,1)$ is
a discount factor. We consider a partially observable setting, where each agent $i$ receives an individual
partial observation $o^i \in\Omega$ according to the observation probability function $O(o^i|s,a^i)$. Each agent
$i$ conditions its individual policy $\pi_i(a^i|\tau^i)$  on its action-observation history $\tau_i \in \mathcal{T} \defeq (\Omega\times \mathcal{A})^*$ 
to jointly maximize team performance. We use $\pmb{\tau} \in \pmb{\mathcal{T}} \defeq \mathcal{T}^n$ to denote the joint action-observation
history.

\subsection{Centralized Training with Decentralized Execution}
Our method adopts the framework of centralized training with decentralized execution (CTDE) \citep{lowe2017multi, foerster2018counterfactual, rashid2018qmix, sunehag2018value}. Agents are trained in a centralized way
and granted access to other agents’ information or the global states during the centralized training
process. However, due to partial observability and communication constraints, each agent makes its
own decision based on its local action-observation history during the decentralized execution phase. One approach to the CTDE framework is value
function factorization.

\section{Methods}
In this section, we introduce our hierarchical learning framework: we first present our latent policy $\pi_m$ detailing each of its objectives as follows: 1) Mutual Information objective: we aim to maximize this quantity in order to encode the information of agents' histories and futures into the latent space, 2) Prediction objective: To predict future segments given agents’ histories and the strategy, 3) Planning Objective: To plan for rewarding strategies based on the future prediction.
Second, we discuss the training procedure of the latent-conditioned policies. And lastly, we explain the model architectures including the method by which we embed the interactions into what is called \textit{relational} strategies. 

\begin{figure}[t]
\begin{center}
\centerline{\includegraphics[width=1\linewidth]{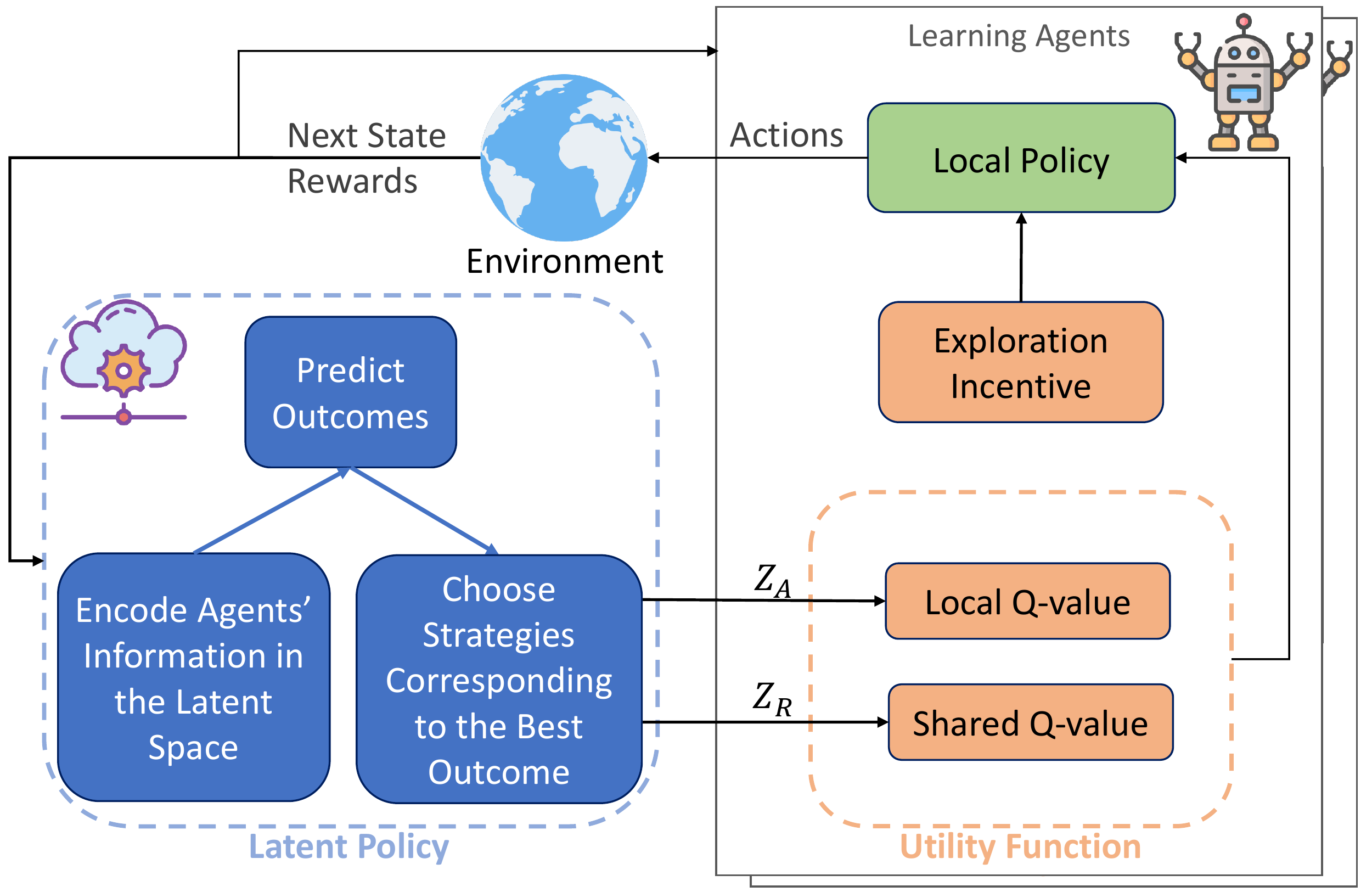}}
\caption{High-level Overview of the structure of  HISMA.}
\label{illus}
\end{center}
\vskip -0.3in
\end{figure}

\subsection{Latent Policy}\label{latent-pol}

In this subsection, we introduce the  desiderata for the latent policy's  objectives. We also show how we can optimize the objectives using a straightforward actor-critic method and how we use the outputs of the policy to learn optimal behaviors for the MARL problem.

Formally, the latent policy $\pi_m:  \pmb{\mathcal{T}}\rightarrow \Delta (\mathcal{Z})$ learns a distribution over the latent  space $\mathcal{Z}$ which we call the "strategies" space. More precisely, a strategy is represented as $z=(z_A^1,z_A^2,\dots,z_A^n,z_R^1,z_R^2,\dots,z_R^n)$ where $z_A^i$ represents individual strategies and $z_R^i$ denotes the relational/social strategies. Agents' local utility functions are provided with $z_R=[z_R^i]_{i=1}^n$ and their corresponding $z_A^i$; we will discuss the former idea in detail in Section (\ref{decomp}). Each choice of a strategy $z$ is sustained for $s$ timesteps (i.e. we divide the horizon into $H$ segments each associated with a strategy, $T=Hs$). To unify notations, let $\tau^i_t=(o^i_1,a^i_1,\dots,o^i_t,a^i_t)$ denote agent $i$'s local trajectories till timestep $t$ ($t\leq T$), and $\zeta_k^i=(o_{ks+1}^i,a_{ks+1}^i,\dots,o_{(k+1)s}^i,a_{(k+1)s}^i)$ the $k$-th segment of agent $i$'s local observations and actions ($k<H$). The primary objective of $\pi_m$ is to learn strategies $z$ such that given agents' histories, it is easy to accurately predict the future behaviors of the agents and harvest the model to plan and make high-rewarding chains of decisions. 
To achieve the former criteria, we introduce the following three objectives:

\subsubsection{Mutual Information Objective} Intuitively, to enable dependency of the agents on the strategies $z$ for inferring the future trajectories (or specifically, the next segment for which the strategy $z$ is learned),  we use an information-theoretic objective to maximize the mutual information between agents' histories $\pmb{\tau}$ and strategies $z$; and their corresponding future segments $\pmb{\zeta}$. More formally, we want to maximize: 
\begin{align*}
     \sum_{k=1}^H \gamma^{ks}\underbrace{\big[I^{\pmb{\pi},\pi_m}(\pmb{\zeta}_k; z_k,\pmb{\tau}_{ks}) +I^{\pmb{\pi},\pi_m}(z_k;\pmb{\zeta}_k,\pmb{\tau}_{ks})\big]}_{{J}^{\pmb{\pi},k}_\text{MI}}
\end{align*}
 By meeting the former objective, we would have satisfied the following criteria: 1) Given $\pmb{\tau}_{ks}$ and $z_k$, the uncertainty of $\pmb{\zeta}_k$ is minimized and vice versa. 2) Given $\pmb{\tau}_{ks}$ and $\pmb{\zeta}_k$, it should be straightforward to infer the strategy $z_k$ that led to $\pmb{\zeta}_k$. To build a better intuition, note that:
\begin{align*}
    {J}^{\pmb{\pi},k}_\text{MI}=& \underbrace{\mathbb{E}_{\pmb{\tau}_{ks}, z_k\sim\pi_m,\pmb{\zeta}_k\sim p(.|z_k,\pmb{\tau}_{ks})}
    \biggl[\log\frac{p\big(\pmb{\zeta}_k|z_k,\pmb{\tau}_{ks}\big)}{p(\pmb{\zeta}_k|\pmb{\tau}_{ks})}\biggl]}_{A}\\
    +& \underbrace{\mathbb{E}_{\pmb{\tau}_{ks}, z_k\sim\pi_m,\pmb{\zeta}_k\sim p(.|z_k,\pmb{\tau}_{ks})}
    \biggl[\log\frac{p\big(z_k|\pmb{\zeta}_k,\pmb{\tau}_{ks}\big)}{p(z_k|\pmb{\tau}_{ks})}\biggl]}_{B}
\end{align*}
 Now, write $A$ as:
\begin{align*}
    A=\mathbb{E}\Biggl[\sum_{t=ks+1}^{(k+1)s} \log & \frac{p(\mathbf{a}_t|\pmb{\tau}_t;z_k)}{p(\mathbf{a}_t|\pmb{\tau}_t)}+\log \frac{p(\mathbf{o}_{t+1}|\mathbf{a}_t,\pmb{\tau}_t;z_k)}{p(\mathbf{o}_{t+1}|\mathbf{a}_t,\pmb{\tau}_t)}\Biggl]
\end{align*}
The first term quantifies the information about how agents behave is gained  when the strategy $z$ is given. The second one encourages observations to be diverse and inferable from the strategy and local trajectories.
\begin{align*}
    B=\mathbb{E}\biggl[\log\frac{p\big(z_k|\pmb{\zeta}_k,\pmb{\tau}_{ks}\big)}{\pi_m(z_k|\pmb{\tau}_{ks})}\biggl]
\end{align*}

This term encourages the latent policy to produce diverse strategies that are identifiable by agents' trajectories. 

\textbf{Information Estimation:} Estimating and maximizing mutual information is often intractable. Drawing inspiration from the literature \citep{bishop:2006:PRML, Wainwright2008GraphicalME, alemi2016deep}, we derive a tractable lower bound of the mutual information which allows differentiability w.r.t the agents' policies' and the latent policy's parameters $(\theta$ and $\omega)$, respectively.
\begin{equation}\label{bound1}
\begin{split}
    A\geq\mathbb{E}\Biggl[\sum_{t=ks+1}^{(k+1)s} &\log  \frac{\sigma(\mathbf{a}_t|\pmb{\tau}_t;z_k)}{p(\mathbf{a}_t|\pmb{\tau}_t)} \cdot \frac{q_\phi(\mathbf{o}_{t+1}|\mathbf{a}_t,\pmb{\tau}_t;z_k)}{p(\mathbf{o}_{t+1}|\mathbf{a}_t,\pmb{\tau}_t)}\Biggl]
    \end{split}
\end{equation}
where $\sigma(\mathbf{a}_t|\pmb{\tau}_t;z)=\prod_i\text{softmax}(Q_i(a_t^i,\tau^i_t;z|\theta^i))$ is the Boltzman operator, and $q_\phi$ is a variational posterior estimator parameterized by $\phi$.
\begin{equation}\label{bound2}
B \geq \mathbb{E}\biggl[\log\frac{q_\xi\big(z_k|\pmb{\zeta}_k,\pmb{\tau}_{ks}\big)}{\pi_m(z_k|\pmb{\tau}_{ks})}\biggl] 
\end{equation}

Also, $q_\xi$ is a variational posterior estimator parameterized by $\xi$. We defer the full derivations to Appendix \ref{math}. Each posterior estimator also minimizes a KL divergence to tighten the bounds in Equations (\ref{bound1},\ref{bound2}), meaning that, $\phi$ also seeks to minimize $D_\phi\defeq\mathbb{E}_{\pmb{\tau},\pmb{a},z}\text{KL}\big(p(.|\mathbf{a},\pmb{\tau};z)||q_\phi(.|\mathbf{a},\pmb{\tau};z)\big)$. The gradient w.r.t $\phi$ becomes: $\nabla_\phi \mathbb{E}_{\pmb{\tau},\pmb{a},\pmb{o},z} \big[\log\frac{p(\pmb{o}|\mathbf{a},\pmb{\tau};z)}{q_\phi(\pmb{o}|\mathbf{a},\pmb{\tau};z)}\big]=-\mathbb{E}_{\pmb{\tau},\pmb{a},\pmb{o},z}[\nabla_\phi\log q_\phi(\pmb{o}|\mathbf{a},\pmb{\tau};z)]$. The case is identical to that of  $q_\xi$ and $D_\xi$. In all, the lower bound we seek to maximize can be written as:
\begin{align*}
    J^{\pmb{\pi},k}_{\text{MI}}&\geq \mathbb{E}\biggl[\log\frac{q_\xi\big(z_k|\pmb{\zeta}_k,\pmb{\tau}_{ks}\big)}{\pi_m(z_k|\pmb{\tau}_{ks})}
    + \sum_{t=ks+1}^{(k+1)s} \log  \frac{\sigma(\mathbf{a}_t|\pmb{\tau}_t;z_k)}{p(\mathbf{a}_t|\pmb{\tau}_t)} \\
    &~~+ \sum_{t=ks+1}^{(k+1)s}\log \frac{q_\phi(\mathbf{o}_{t+1}|\mathbf{a}_t,\pmb{\tau}_t;z_k)}{p(\mathbf{o}_{t+1}|\mathbf{a}_t,\pmb{\tau}_t)}\biggl] \defeq {J}^k_\text{MI} (\xi,\phi,\omega,\theta)
\end{align*}

\subsubsection{Reconstruction  Error Objective}
The former objective ensures that the latent space is learned and dependency is established, we thus propose to learn a predictive model that outputs future segments when given agents' histories and strategies. Specifically, let $F_\eta$ be the segment predictive model parameterized by $\eta$. $F_\eta$ takes $\pmb{\tau}_{ks}$ and $z_k$ as inputs and returns the expected segment of agents' trajectories $\pmb{\zeta}_k$ that follows the history $\pmb{\tau}_{ks}$ and the strategies $z_k$. For training,  we sample $\{\pmb{\tau}_{k_ls},\pmb{\zeta}_{k_l}\}_1^{l}\sim\mathcal{D}$. We then, for the sake of compactness, stack these vectors into matrices $\pmb{T}, \pmb{\Psi}$, respectively. We use the Frobenius norm, and write the reconstruction error  as:
\begin{align*}
    &\min_{\eta,\omega}~ \mathbb{E}_{\pmb{\mathrm{Z}}\sim\pi_m(\pmb T)} \biggl[\norm{\pmb{\Psi}-{F}_{\eta}(\pmb{T},\pmb{\mathrm{Z}})}_F\biggl]
\end{align*}
We call the former objective the prediction error objective and denote it by $J_e(\eta,\omega)$. This error model
enables the latent policy to reconstruct the ground truth trajectories accurately.

\subsubsection{Forecasted Trajectories Rewards Objective}
Lastly, since the latent space is established and consistent with agents' information, it is rather natural to learn to output sequences of strategies such that their corresponding predicted future segments yield the maximum returns possible. More formally, if we denote the sum of agents' shared extrinsic rewards over a segment $\pmb{\zeta}_k$ by $R(\pmb{\zeta}_k)$, then the objective $J_m(\omega)$ can be written as follows: 
\begin{align*}
    \max_{\omega}~& \mathbb{E}_{\hat{\pmb{\tau}}}\Biggl[\sum_{k=1}^{H}\gamma^{ks} R(\hat{\pmb{\zeta}}_k)\Biggl]\\
    \text{s.t.}~~& \hat{\pmb{\tau}}_{ks}=(\hat{\pmb{\zeta}}_1,\dots,\hat{\pmb{\zeta}}_k), z_k \sim \pi_m(\hat{\pmb{\tau}}_{ks})\\
    &    \hat{\pmb{\zeta}}_{k} = F_\eta(\hat{\pmb{\tau}}_{ks},z_k) ~\text{or}~ \hat{\pmb{\zeta}}_{k} \sim p_{\phi,\theta}(.|z_k,\hat{\pmb{\tau}}_{ks})
\end{align*}

In practice, we optimize $J_\text{MI}\defeq\sum_k\gamma^{ks}J_\text{MI}^k$ and $J_m$ using an on-policy (online) policy optimization method (such as PPO) over the reward field
\begin{align*}
    r(\pmb{\tau}&_{ks},z_k)=\lambda_m R(\pmb{\zeta_k})+
    \lambda_\text{MI}\log\frac{q_\xi\big(z_k|\pmb{\zeta}_k,\pmb{\tau}_{ks}\big)}{\pi_m(z_k|\pmb{\tau}_{ks})} \\
    &~+ \lambda_\text{MI}\sum_{t=ks+1}^{(k+1)s}\biggl[\log  \frac{\sigma(\mathbf{a}_t|\pmb{\tau}_t;z_k)}{p(\mathbf{a}_t|\pmb{\tau}_t)}\cdot \frac{q_\phi(\mathbf{o}_{t+1}|\mathbf{a}_t,\pmb{\tau}_t;z_k)}{p(\mathbf{o}_{t+1}|\mathbf{a}_t,\pmb{\tau}_t)}\biggl]
\end{align*}

\subsection{Value Decomposition for Relational \& Individual Strategies}\label{decomp}

Previously, we introduced the three objectives of the latent policy for learning strategies by expecting the future behaviors of the learning agents and planning according to the most rewarding outcome. The strategies are categorized into individual strategies (i.e. $z_A^i$) and relational strategies (i.e. $z_R^i$), where the former tends to control the agent-specific behaviors while the latter takes into account the team relations and plans accordingly. We also introduced a method to disentangling both quantities by feeding $z_R$ into a modified Graph Attention Network to learn the importance of the information received from other agents  which in turn characterizes the relations within the team. However, the local $Q$-functions do not have enough capacity to represent both agent-specific and team-related strategies. This encourages us to introduce a shared $Q$-function $Q_R$ that conditions on $z_R$ and is added to an agent-specific utility  $Q_A^i$. More specifically, $Q_A^i$ receives agent $i$'s individual strategy $z_A^i$. In all, we present agent $i$'s utility functions, $Q_i$, as: 
\begin{align*}
    Q_i(a^i,\tau^i;z|\theta^i)=Q_A^i(a^i,\tau^i;z_A^i)+Q_R(a^i,\tau^i;z_R)
\end{align*}

\textbf{Data Gathering Procedure}

To incentivize agents to collect diverse data that help compute accurate estimates of the aforementioned objectives, we regularize the returns objective of the agents by:    
\begin{align*}
    \mathbb{E}_{\pmb{\tau},z_{1:H}}\bigg[\sum_k\alpha\gamma^{ks}\norm{\pmb{\zeta}_k-F_\eta(\pmb{\tau}_{ks},z_k)}_2+\beta\gamma^{ks} h(J^k_\text{MI})\bigg]
\end{align*}

where $\alpha,\beta>0$, $h$ is a  positive and  strictly \textbf{decreasing} function, and the quantity in the expectation is used as a scalar reward to the agents over a complete trajectory. The primary intuition behind the formula is that it quantifies \textit{surprise}, since higher values mean minimal mutual information w.r.t the given trajectory and greater residual error thus storing these diverse data in the replay buffer ensures collecting sufficient information needed for prediction and learning. 

\textbf{Learning Objectives in a Nutshell}

We follow the literature by passing our modified utility functions through a QMIX-style mixing network to obtain the global $Q$-function to be later used to compute the conventional TD error $\mathcal{L}_\text{TD}$.
The overall learning objectives become: (See Algorithm \ref{alg})
\begin{equation}\label{all-objs}
\begin{split}
    \max_{\theta,\omega,\phi,\xi,\eta} &\lambda_m J_m(\omega)-\lambda_e J_e(\eta,\omega)\\
    &- \lambda_\text{TD} \mathcal{L}_\text{TD}(\theta)+\lambda_\text{MI} \sum_k \gamma^{ks} J_\text{MI}^k(\xi,\phi,\omega,\theta)
    \end{split}
\end{equation}

\subsection{Model Architectures} \textbf{Latent Policy.} As described in Figure (\ref{illus-big}) {left}, the first block of the latent policy computes a conditional distribution $\pi_m(z|\pmb{\tau}_t)$ where $z=(z_A^1,z_A^2,\dots,z_A^n,z_R^1,z_R^2,\dots,z_R^n)$. Specifically, we concatenate $\pmb{\tau}_t=(\tau^1_t,\tau^2_t,\dots,\tau^n_t)$ and use a recurrent neural network (RNN)  to compute the features of agents' local trajectories: $\text{RNN}(\pmb{\tau}_t)=X\in\mathbb{R}^{M\times n}$ where $M$ is the feature dimension. We then assume a complete graph $G$ for the agents, and use a graph convolutional network to improve features extraction: $Y=\text{GCN}(G,X)$ where $Y\in\mathbb{R}^{M\times n}$. Next, we assume that $z_A^i,z_R^i$ follow  multivariate Gaussian distributions, i.e. $z_A^i\sim \mathcal{N}(\pmb{\mu}_A^i,\pmb{\sigma}_A^i)$ and $\tilde{z}_R^i\sim \mathcal{N}(\pmb{\mu}_R^i,\pmb{\sigma}_R^i)$. The distributions parameters are predicted by shared  multi-layer perceptions (MLP) by passing the corresponding column $Y_i$:
\begin{align*}
    &\pmb{\mu}_A^i,\pmb{\sigma}_A^i=F_{Z_A}(Y_i;\omega_{Z_A})\\
    &\pmb{\mu}_R^i,\pmb{\sigma}_R^i=F_{Z_R}(Y_i;\omega_{Z_R})
\end{align*}

As  mentioned, the latent variable $z_R$ models the relations between agents by encoding the information necessary for achieving sufficient coordination, and is communicated to a shared network $Q_R$ that is later added to each agent's local utility function (See Section (\ref{decomp})). We propose to model the relations by incorporating a modified graph attention network (GAT) and assuming a complete graph $G$ for the agents. Specifically, we use agents' local histories to compute the attention weights on edges and update node features with the latent variables $\tilde{z}_R^i$. More formally, we write:
\begin{align*}
    \alpha_{ij}=\text{softmax}_j \langle F_d(\tau^i_t), F_s(\tau^j_t)\rangle
\end{align*}
where $F_s, F_d$ are two GRU+MLP networks for source and destination
node features respectively, $\langle \cdot ,\cdot \rangle$ denotes the dot product, and $\{\alpha_{ij}\}$s are the attention weights used for the graph computation.

Consequently, Given the weights $\alpha_{ij}$ and  the latent variables $\tilde{z}_R$, we perform one round of message passing to extract the features that characterize the relations between the agents:
\begin{align*}
    z_R^i=\mathrm{S}\biggl (\sum_j \alpha_{ij} F_g \big([\tilde{z}_R^i,\tilde{z}_R^j]\big)\biggl)
\end{align*}
where $\mathrm{S}(.)$ is the sigmoid function, and $F_g$ is an MLP.

\textbf{Prediction Model \& Posterior Estimators.} \label{F-arch}

The estimators $F_\eta$ and $q_\phi$ have a similar architecture: we seek to model a joint segment of observations and actions in both, thus it is appropriate to use an auto-regressive model with dilated convolutions for each \citep{oord2016wavenet,oord2016conditional,mishra2017prediction} and an activation function of the form:
\begin{align*}
    \mathrm{S}(W_{g,k}^i*x^i + (V^i_{g,k})^T z^i) \odot \tanh (W_{f,k}^i*x^i + (V^i_{f,k})^T z^i)
\end{align*}

where $\mathrm{S}$ is the sigmoid function; $k$ is the layer index; $x^i,z^i=[z^i_A,z^i_R]$ are the input to the layer  and the latent strategy that correspond to agent $i$. $W,V$ are network weights that comprise the parameters  the model. Also, $\odot$ and $*$ denote element-wise multiplication and convolution operators, respectively; $f,g$ denote filter and gate, respectively.

The posterior estimator $q_\xi(.|\pmb{\zeta},\pmb{\tau})$ has a similar architecture to that of $\pi_m(.|\pmb{\tau})$, except for using a backward RNN to process $\pmb{\zeta}$. Thus, after processing the sequences $\pmb{\tau},\pmb{\zeta}$, we concatenate the outputs of their corresponding RNNs and pass them through a GCN with a complete graph for the agents, and lastly, each column of the resulting matrix is used as an input to shared MLPs to compute the parameters of the distribution $q_\xi$.


\begin{figure*}[ht]
\begin{center}
\centerline{\includegraphics[width=\linewidth]{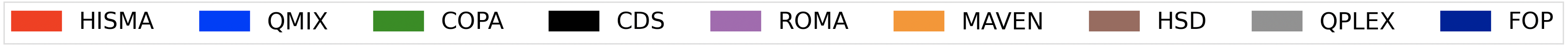}}
\subfigure[2c\_vs\_64zg \textbf{Hard}]{
        \includegraphics[width=0.2333\textwidth]{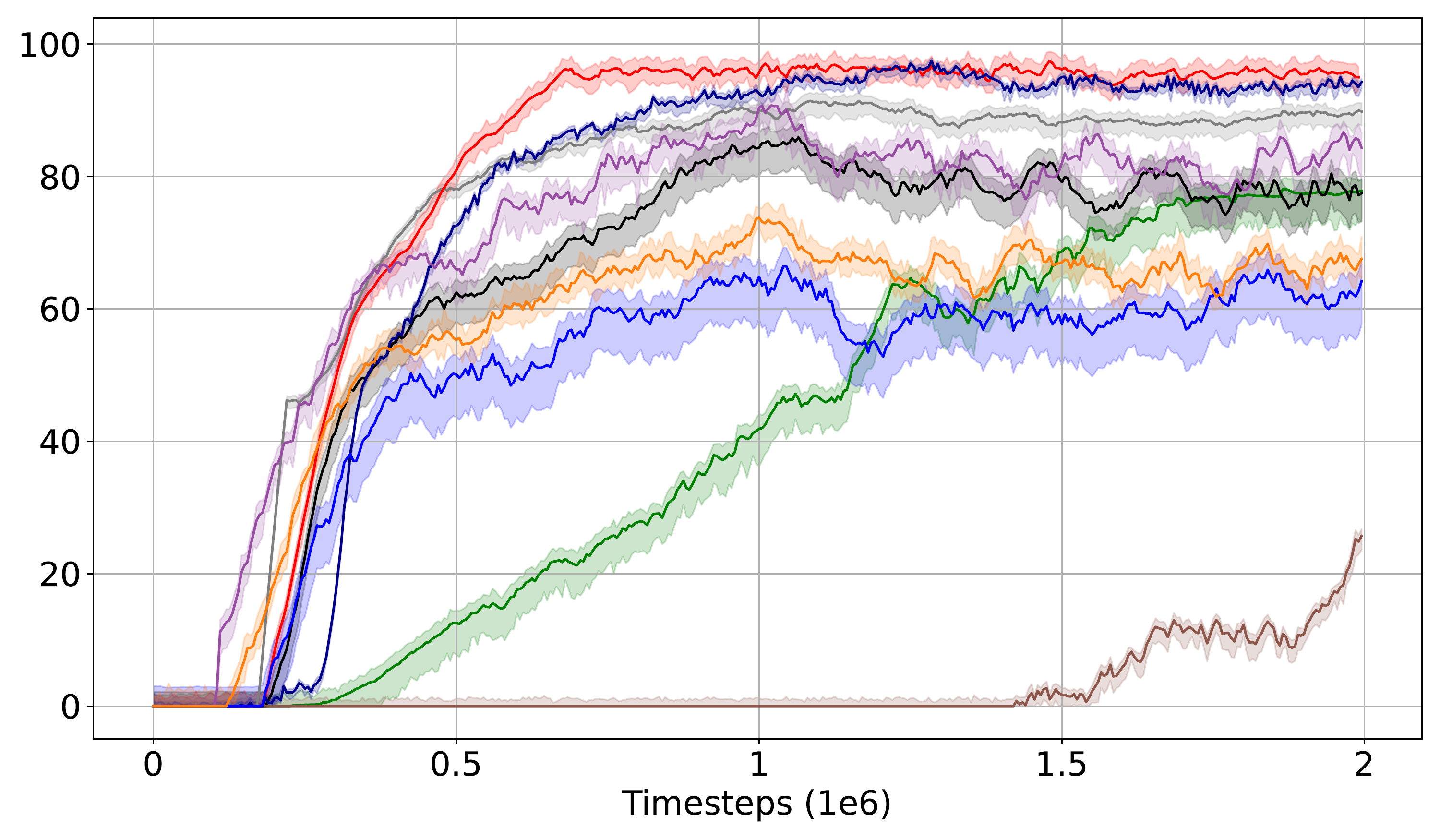}
    }
\subfigure[6h\_vs\_8z \textbf{SuperHard}]{
        \includegraphics[width=0.2333\textwidth]{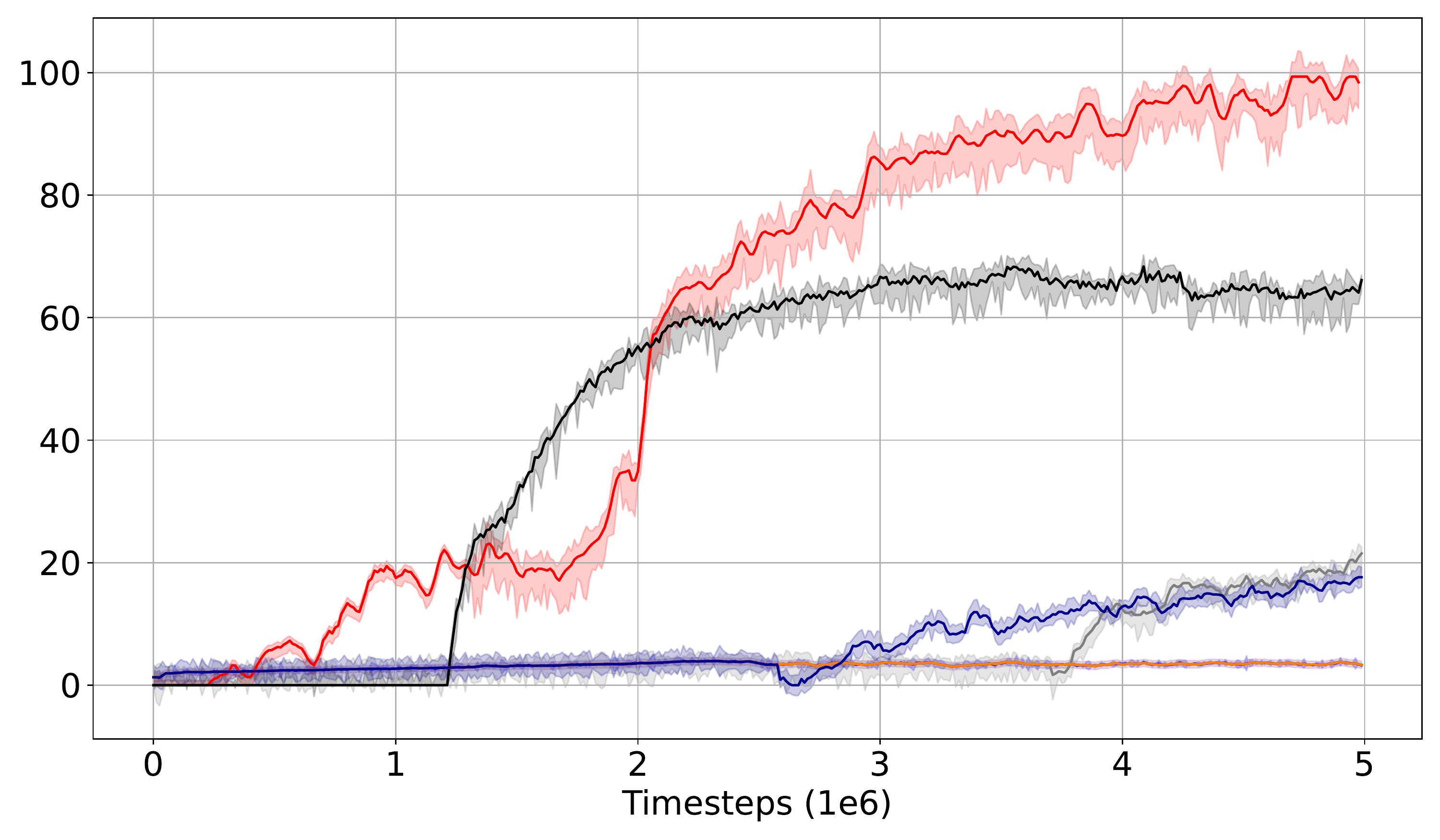}
    }
\subfigure[Corridor \textbf{SuperHard}]{
        \includegraphics[width=0.2333\textwidth]{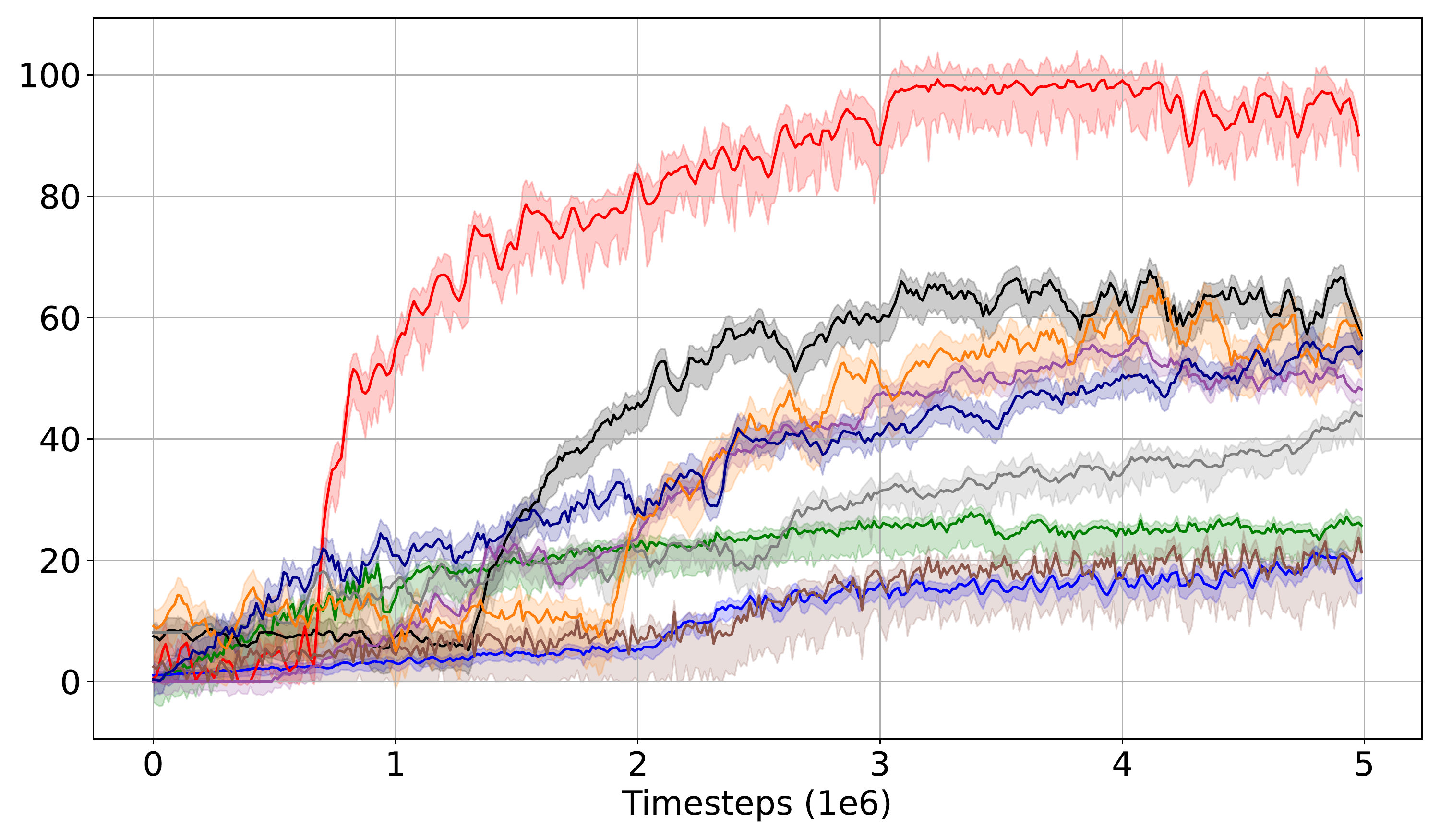}
    }
\subfigure[3s5z\_vs\_3s6z \textbf{SuperHard}]{
        \includegraphics[width=0.2333\textwidth]{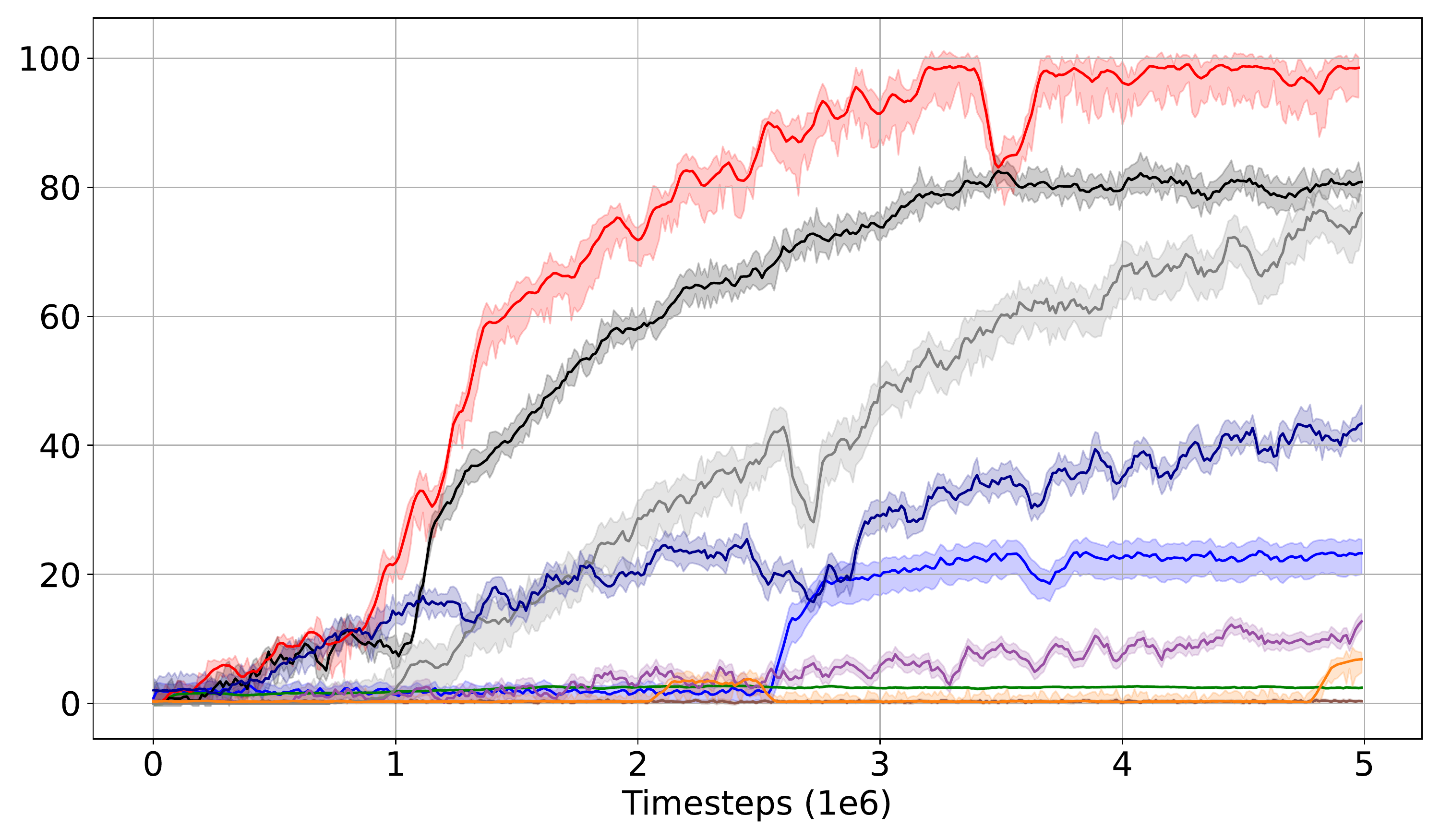}
    }
\subfigure[MMM2 \textbf{SuperHard}]{
        \includegraphics[width=0.2333\textwidth]{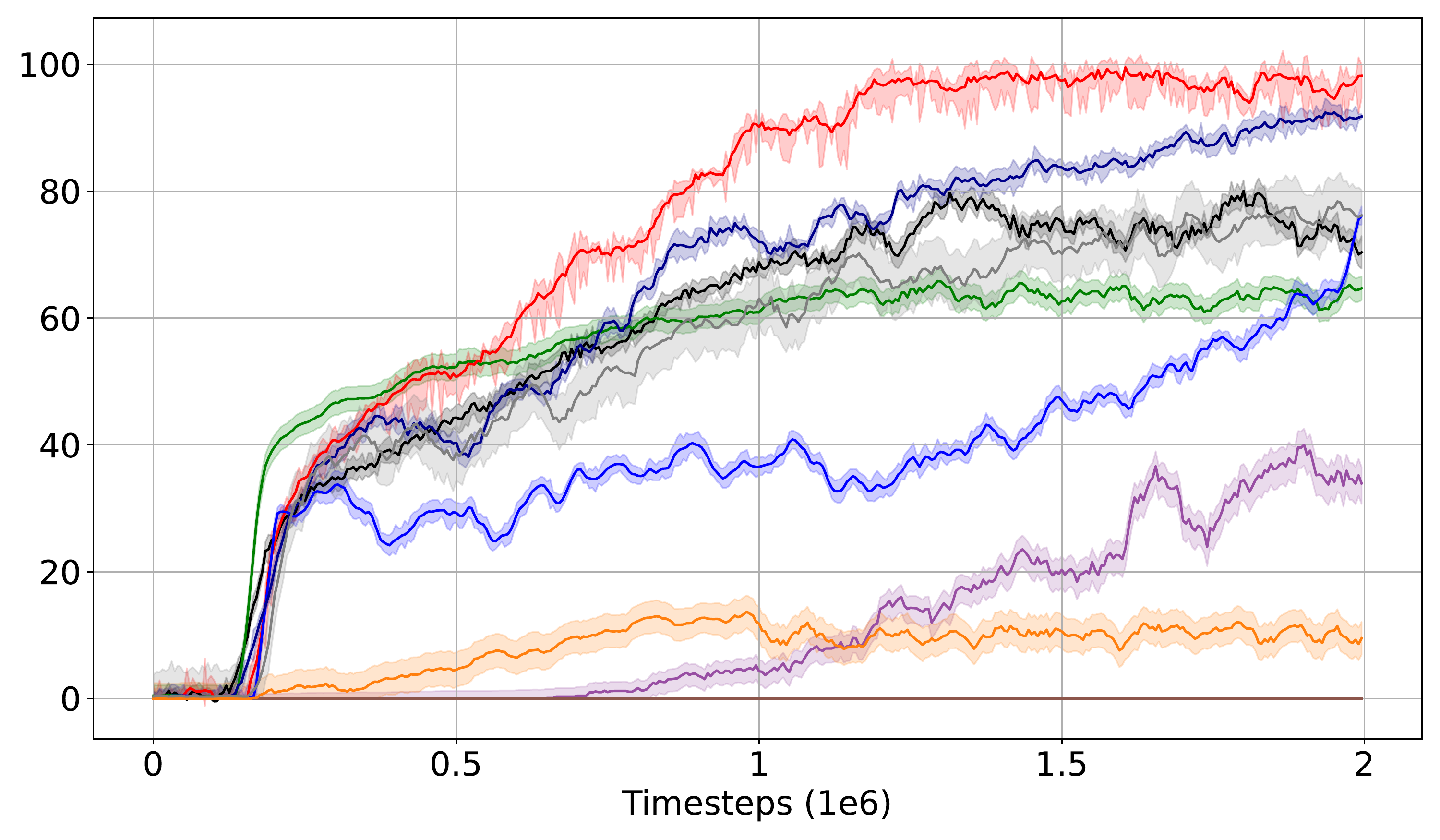}
    }
\subfigure[27m\_vs\_30m \textbf{SuperHard}]{
        \includegraphics[width=0.2333\textwidth]{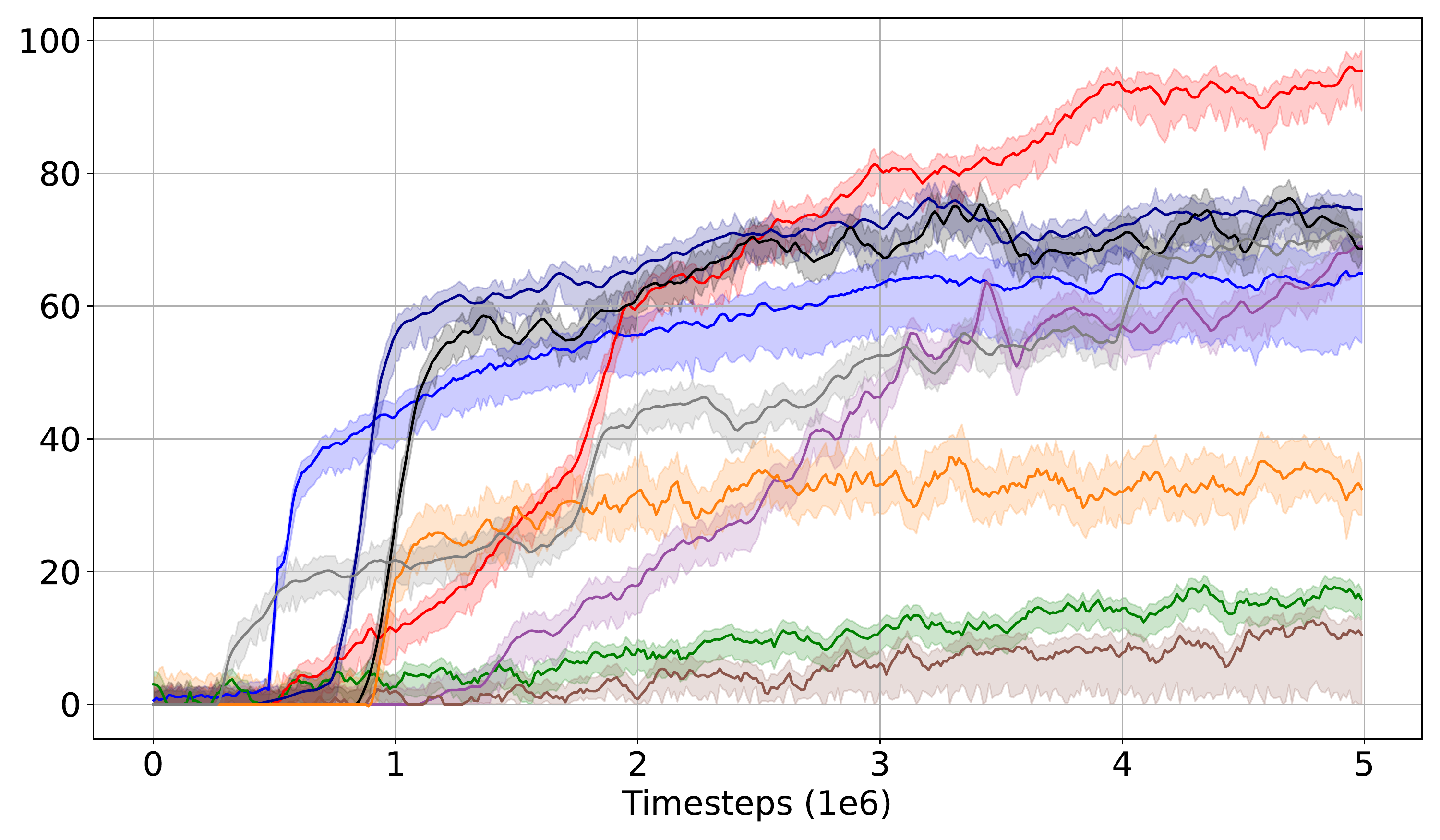}
    }
    \subfigure[1c3s8z vs 1c3s9z \textbf{SuperHard}]{
        \includegraphics[width=0.2333\textwidth]{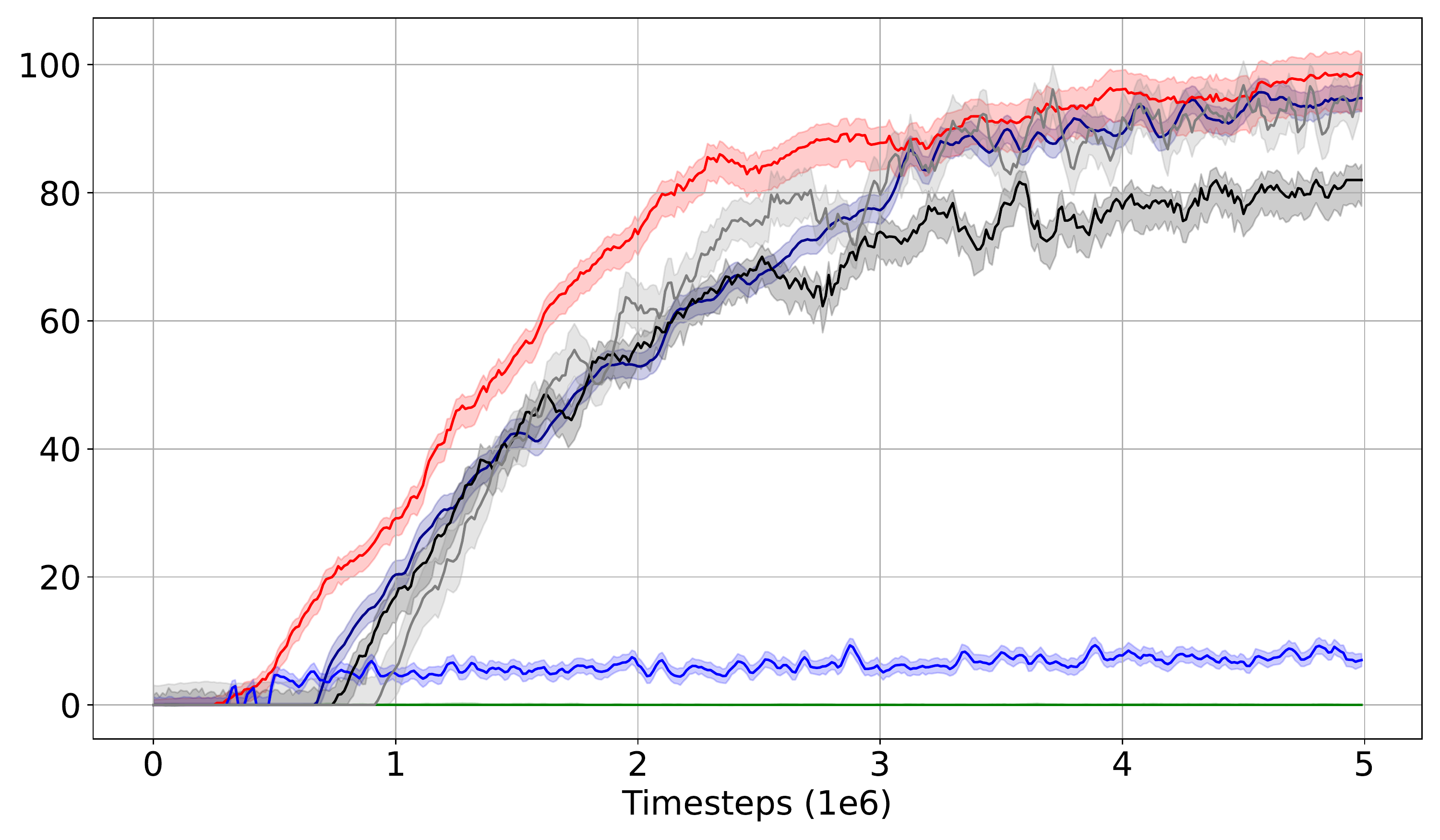}
    }
    \subfigure[5s10z \textbf{SuperHard}]{
        \includegraphics[width=0.2333\textwidth]{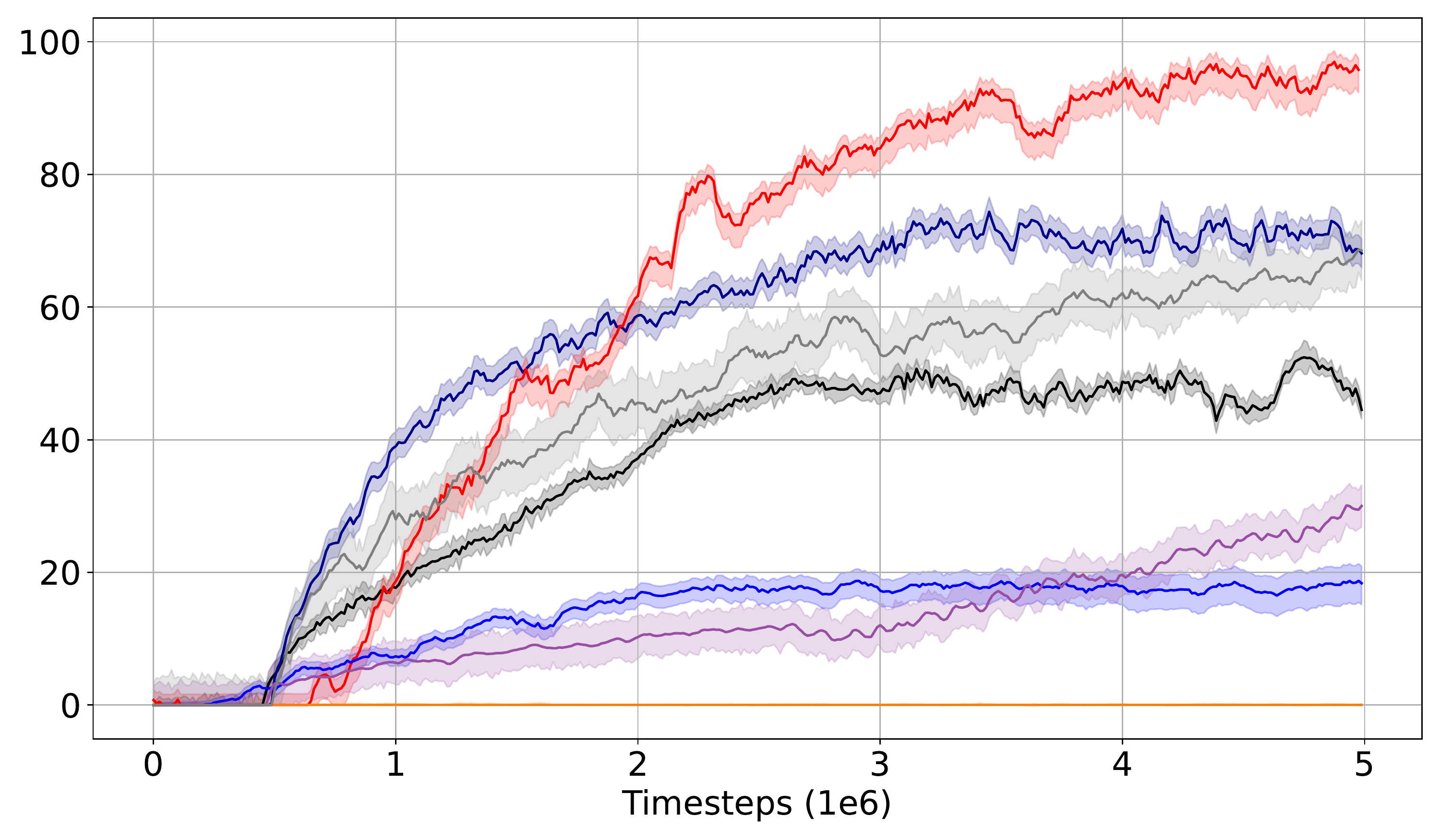}
    }
\caption{The median test win \%  of various methods across the SMAC scenarios.}
\label{smac}
\end{center}
\vskip -0.2in
\end{figure*}

\section{Empirical Evaluation \& Analysis}\label{experiments}
The goals of our experiments are to: a) Verify that our learned strategies improve the performance of multi-agent reinforcement learning on a comprehensive set of challenges against the current state-of-the-art methods; b) perform ablations to examine which particular components of the proposed framework are important for  performance gains; c) articulate and visualize the strategies learned by our method.\footnote{All experiments are run for 12 random seeds. Hyperparameters and computational resources are discussed in Appendix \ref{hyper}.  Code base is also provided.}

\begin{figure*}[ht]
\begin{center}
\centerline{\includegraphics[width=\linewidth]{additional_pic/LEGENDS.png}}
\subfigure[SuperHard Counter Attack]{
        \includegraphics[width=0.233\textwidth]{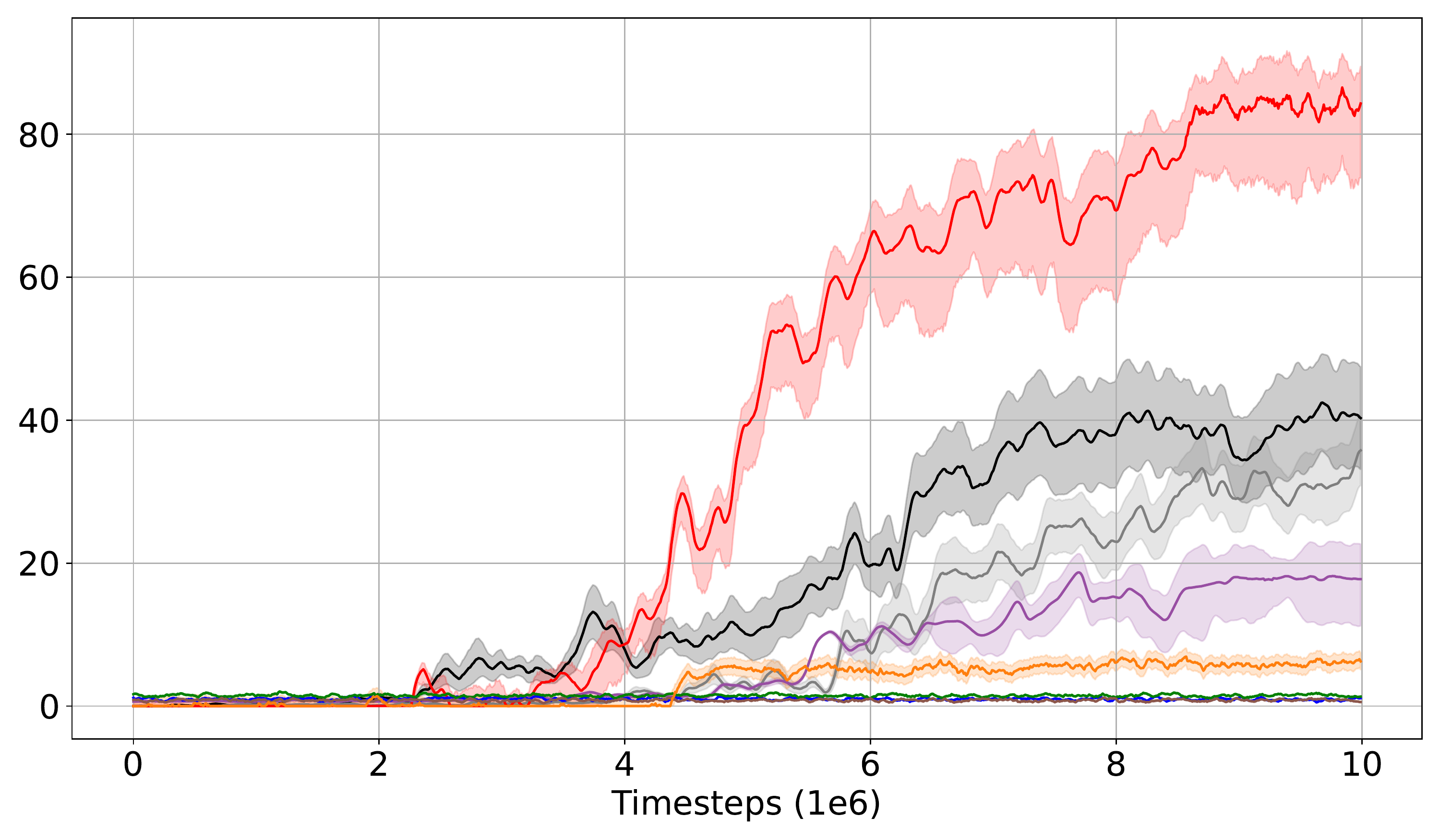}
    }
\subfigure[Run to Score with Keeper]{
        \includegraphics[width=0.233\textwidth]{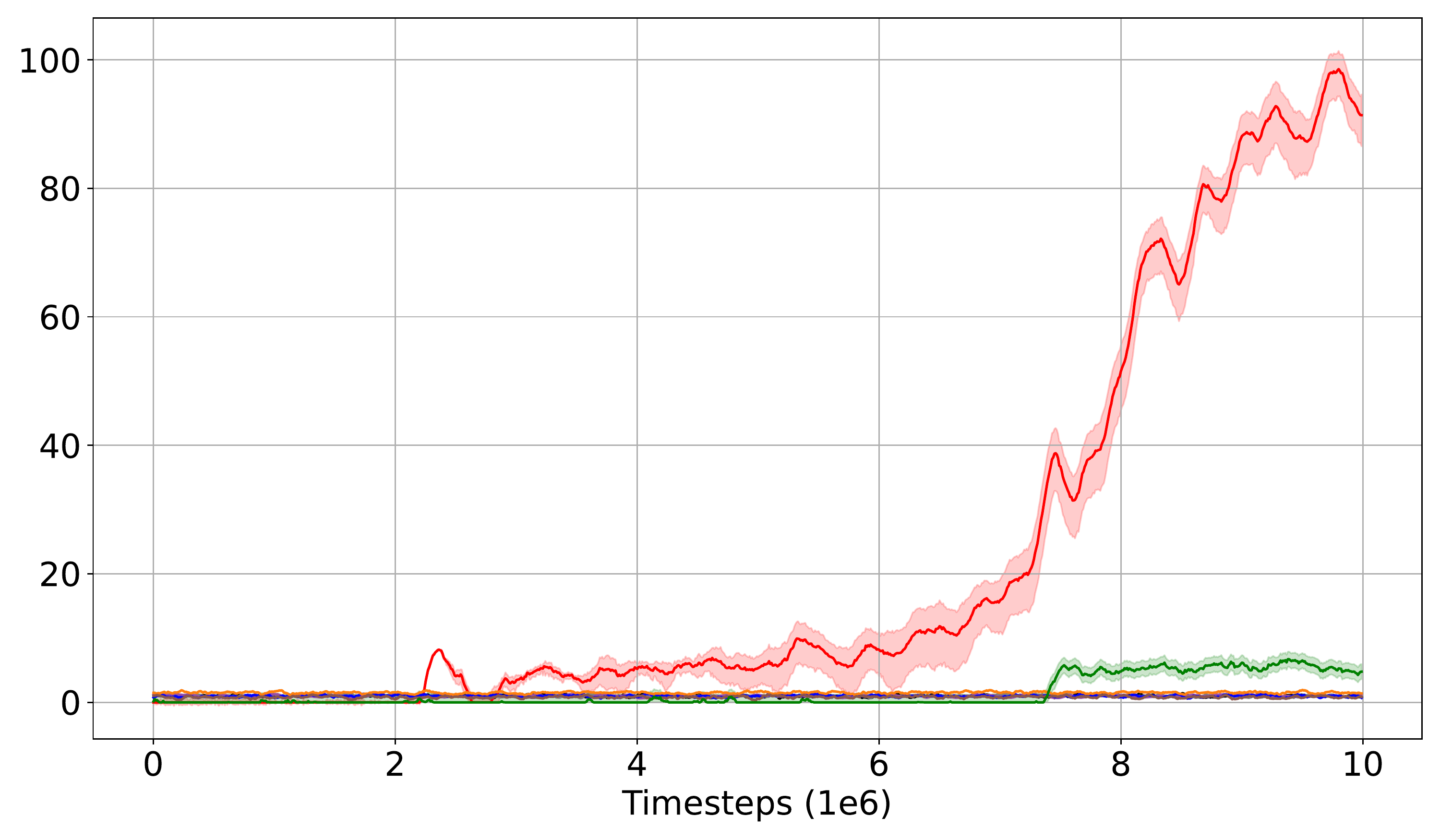}
    }
\subfigure[3 vs 1 with Keeper]{
        \includegraphics[width=0.233\textwidth]{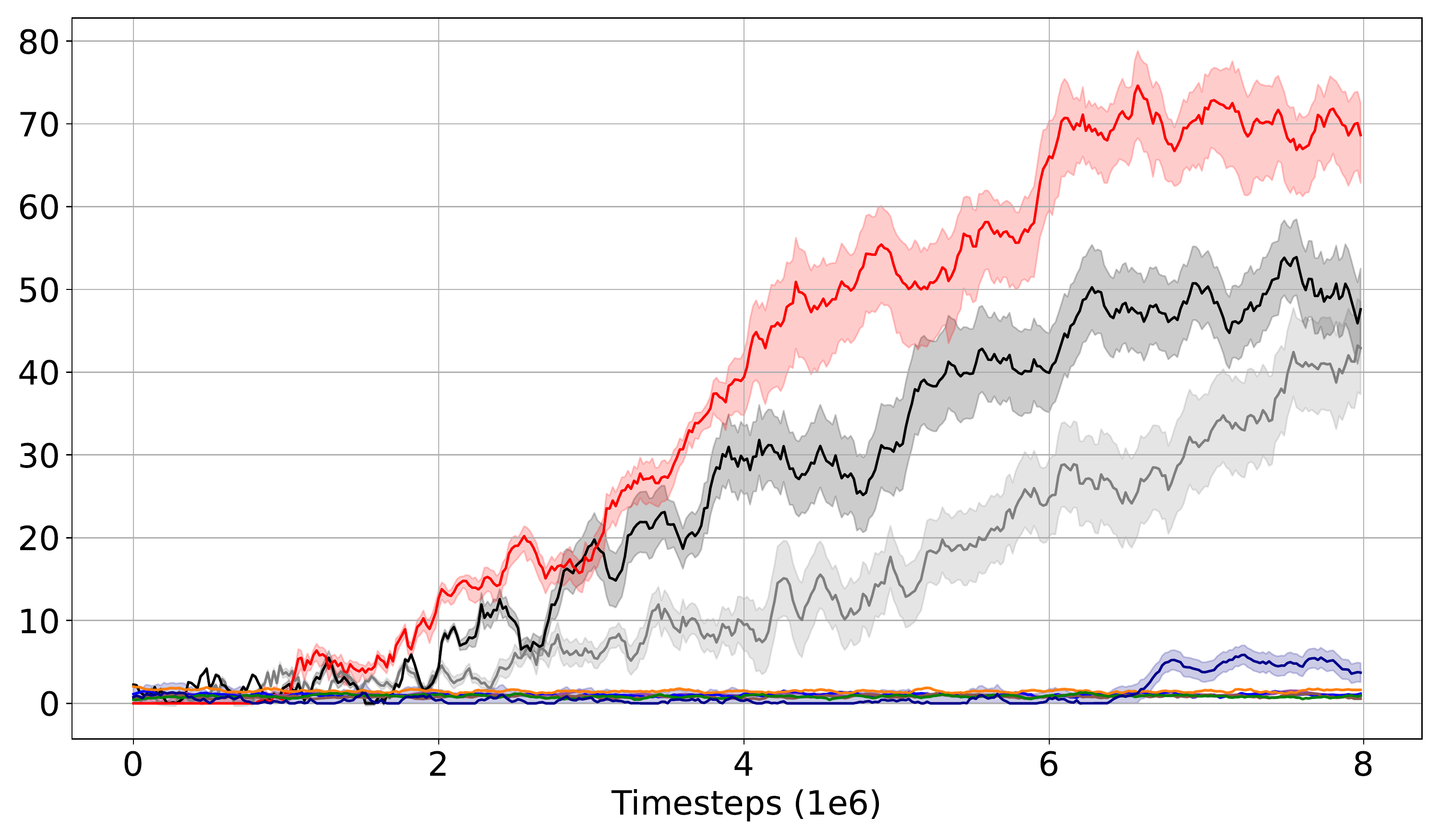}
    }
\subfigure[Corner]{
        \includegraphics[width=0.233\textwidth]{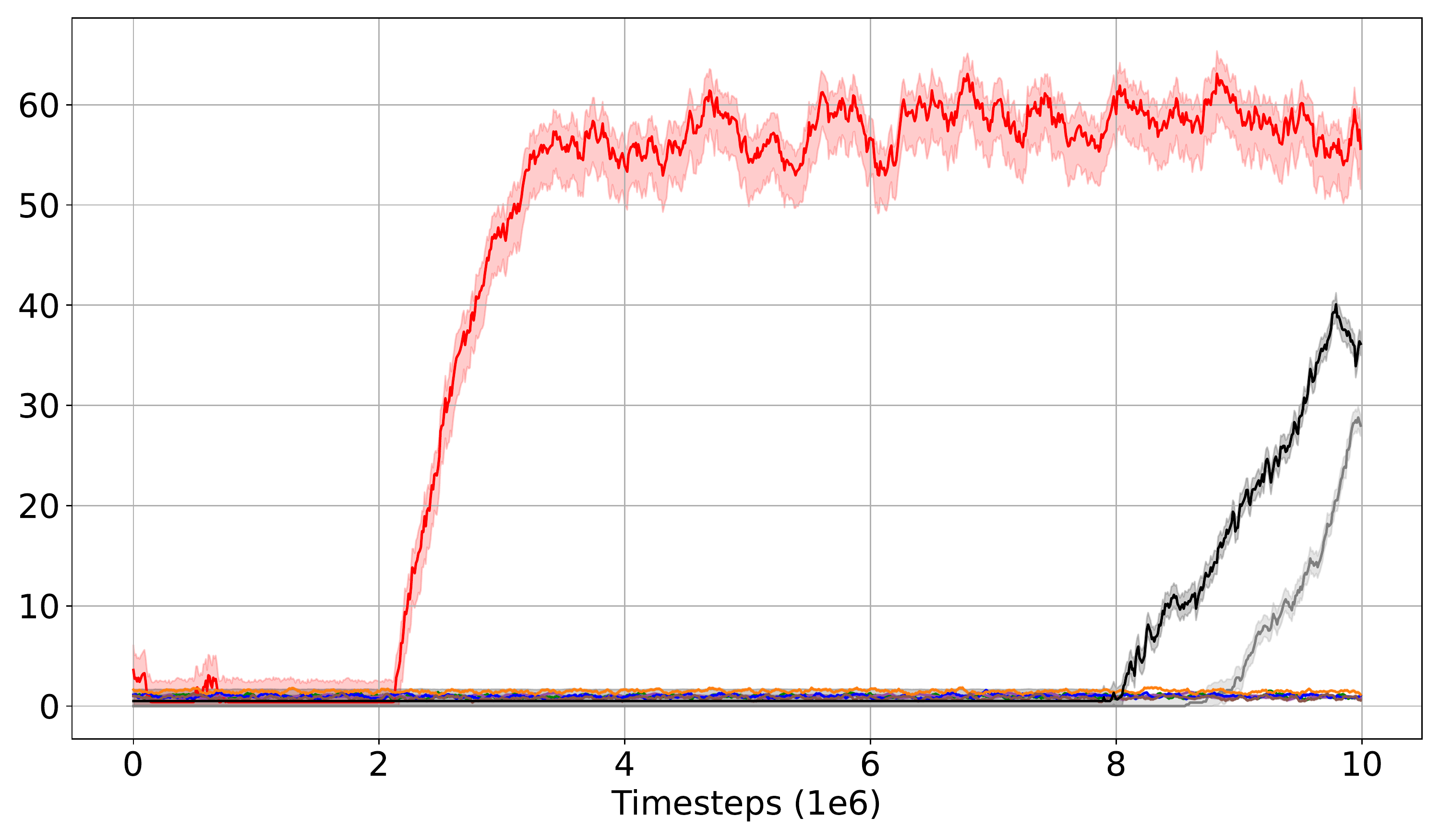}
    }
\caption{The median test win \%  of various methods across the GRF tasks.}
\label{grf}
\end{center}
\vskip -0.2in
\end{figure*}

\textbf{Baselines.} We benchmark our approach on extremely challenging tasks, and demonstrate the superior performance of our methods against the current state-of-the-art algorithms using their open-source implementations:  QPLEX: demonstrated the best performance among value-based MARL algorithms \citep{wang2021qplex}; FOP: an actor-critic method of factorized individual policies with provable convergence to the global optimum \citep{pmlr-v139-zhang21m}; CDS: a powerful method that promotes diversity within a shared team and enlarges the representational capacity of utility functions \citep{chenghao2021celebrating};  QMIX \citep{rashid2018qmix}. \textit{Hierarchical MARL baselines}: COPA: Coach Player MARL with a hierarchy to learn strategies \citep{pmlr-v139-liu21m};    ROMA: Role-based MARL  \citep{wang2020roma}; HSD: Hierarchical Cooperative MARL with Skills Discovery \citep{yang2020hierarchical}; MAVEN: Multi-Agent Variational Exploration with a hierarchical choice of behavior modes \citep{mahajan2019maven}.
We further test against two very recent SOTA methods: EMC \citep{zheng2021episodic} and RODE \citep{wang2021rode}, and demonstrate the consistent superiority of HISMA. However, we omit their learning curves from Fig. (\ref{smac},\ref{grf}) for visual clarity and defer the reports to Fig. (\ref{more-smac}).

\subsection{Performance on StarCraft II}

The StarCraft II unit micromanagement task is considered as one of the most challenging cooperative multi-agent testbeds
for its high degree of control complexity and environmental stochasticity.  SC II (or SMAC) provides a rich set of units each with diverse actions, allowing for extremely complex cooperative behaviors among agents. We thus evaluate our method on several SC II micromanagement tasks from the SMAC benchmark  \citep{samvelyan2019starcraft} We consider the $7$ \textbf{Super Hard} battle maps:  3s5z\_vs\_3s6z, Corridor,  27m\_vs\_30m, 6h\_vs\_8z, MMM$2$, 5s10z, and 1c3s8z\_vs\_1c3s9z along with a \textbf{Hard} scenario: 2c\_vs\_64zg.\footnote{We further describe the details  of each  scenario in Appendix \ref{smac-app} \& Table (\ref{smacenvs}).}

Figure (\ref{smac}) shows that our method yields substantially better results than all SOTA MARL methods, sometimes with a significant gap up to $40\%$ in win rate. Also, note that HISMA consistently learns to solve the super hard tasks with almost $100\%$ win rate within only 5 million steps, which is unprecedented compared to other MARL algorithms.

\subsection{Performance on Google Research Football}
We test HISMA on the set of Google Research Football (GRF) \citep{Kurach2020GoogleRF} challenging tasks; two main challenges are manifested in this set: 1) \textbf{Extreme stochasticity}: GRF offers several types of randomness, for example, the same shot may lead to several different outcomes. 2) \textbf{Sparse rewards}: In full-game mode, environmental reward only occurs when a goal is scored, meaning that agents need to coordinate their behaviors to discover complex tactics that lead to scoring goals. Precisely, a +1 reward is given when our team scores, and a penalty of -1 when a goal is received during the 3000 time-steps of the game. In GRF tasks, the reward is also sparse and only given at the end of the task as +100 if agents succeed and -1 otherwise. We set the difficulty level to 1 (hardest), speeding up the bot reaction time and decision making. All baselines are tested with the same level of difficulty.

We also show that HISMA can ace the multi-agent Google Research Football full game by comparing its performance (Win rate, Goal Difference, and TrueSkill rating \citep{herbrich2006trueskill}) against SOTA MARL algorithms. Results after 72 million steps of training are reported in Table (\ref{grf-full}). To the best of our knowledge, HISMA is the first multi-agent algorithm that achieves a win rate of $95.6\%$ with a skill rating of $31.87 \pm 4.33$ thus significantly outperforming all the powerful methods in the field.

To further verify the learning efficiency of HISMA, we show the performance comparison against baselines on the tasks: {SuperHard Counter Attack, Run to Score with Keeper, 3 versus 1 with Keeper, and Corner Kick}, and report the results in Figure (\ref{grf}). As a side note, in the SuperHard Counter Attack, the agents combat against eleven players making this version much harder than the 4-player version of the task. CDS generally shows good performance when compared to other baselines since it incentivizes more efficient exploration. However, our method makes use of the structure of the latent strategies  to ensure individuality and efficient coordination simultaneously which explains its excellent performance and learning speed. Also, by optimizing the three objectives, the latent policy facilitates the dynamic property of behaviors so that agents can switch to different strategies when facing different scenarios.

\begin{table}[h!]
\caption{Evaluation results of different algorithms on the multi-agent Google Research Football full game after 72 million steps.}
\label{grf-full}
\begin{center}
\begin{small}
\begin{sc}
\begin{tabular}{lcccr}
\toprule
Method & Win Rate  & Goal Diff. & TrueSkill$^{\text{TM}}$\\
\midrule
\textbf{HISMA}    & \textbf{95.6} & \textbf{3.914} & \textbf{31.87 $\pm$ 4.33} \\
EMC   & 90.0& 3.037&    26.75$\pm$3.08     \\
CDS   & 88.5& 2.879&    24.47$\pm$4.20     \\ RODE   & 87.7& 2.250& 24.03$\pm$2.15 \\
FOP   & 81.1& 2.365& 23.16$\pm$2.68 \\
QPLEX   & 78.2& 2.522& 22.34$\pm$1.19 \\
ROMA     & 73.3& 1.746& 21.87$\pm$3.39\\
COPA    & 67.9& 0.893& 20.13$\pm$5.65 \\
MAVEN      & 68.4& 0.916& 19.27$\pm$ 5.02        \\
QMIX & 68.7 & 1.043& 19.09$\pm$2.75\\
HSD   & 44.9& -0.089& ~8.57$\pm$3.79\\
\bottomrule
\end{tabular}
\end{sc}
\end{small}
\end{center}
\vskip -0.3in
\end{table}

\subsection{Further Analysis}

\begin{figure*}[t]
\begin{center}
\subfigure[Ablations of Objectives on Corridor]{
        \includegraphics[width=0.315\linewidth]{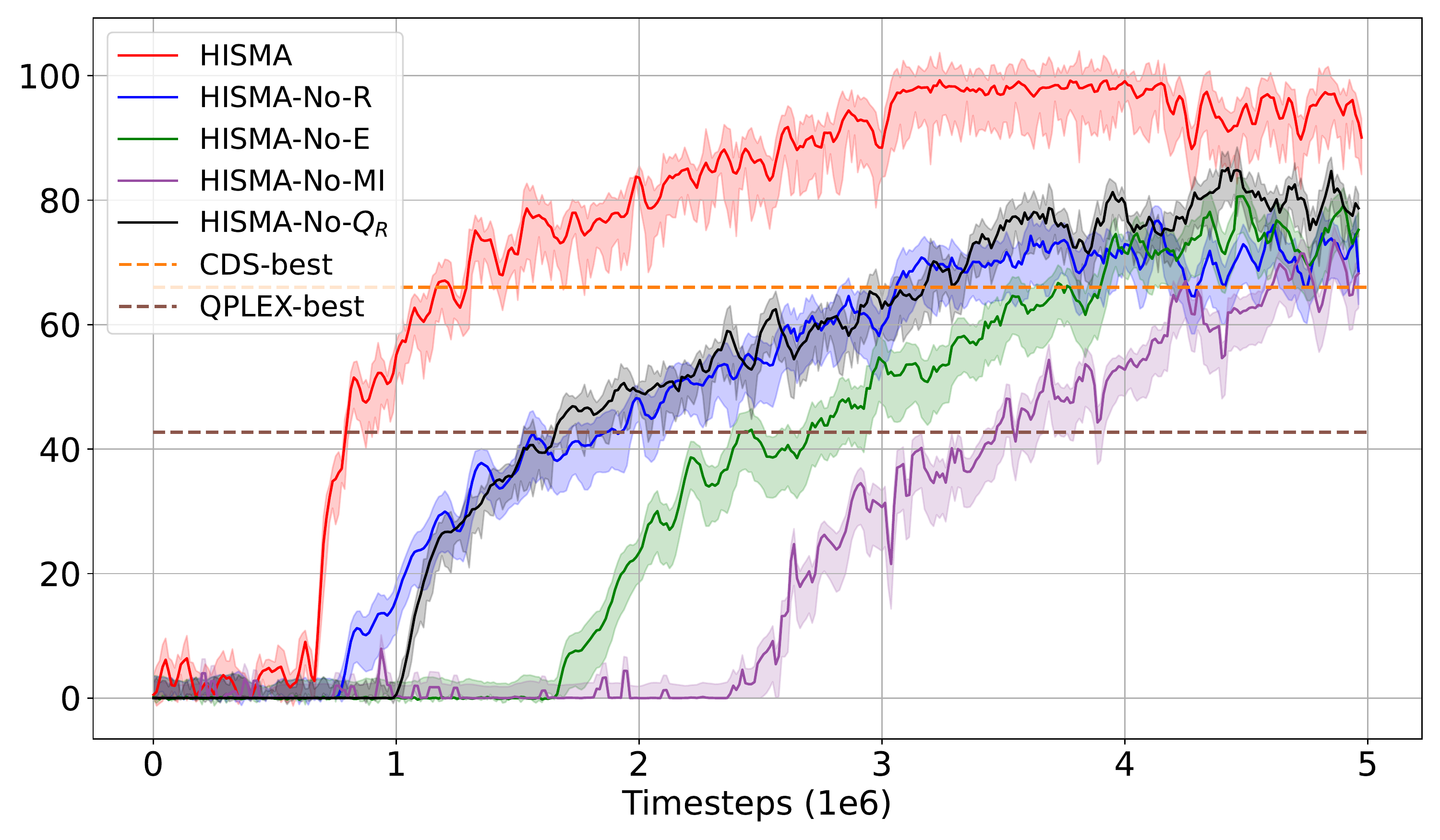}}
\subfigure[Analysis of $s$  on 6h\_vs\_8z\label{abl-b}]{
        \includegraphics[width=0.315\linewidth]{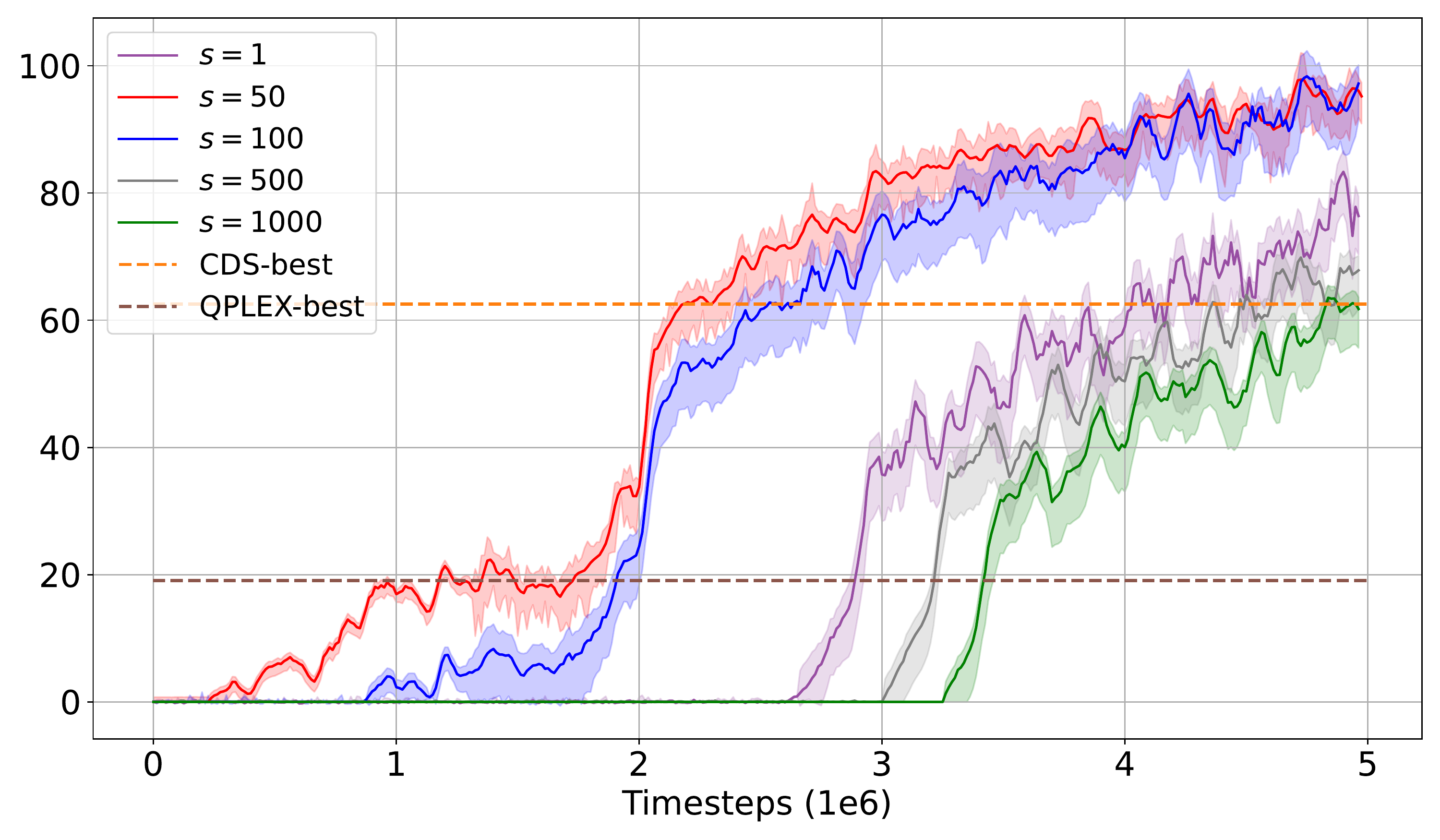}}
\subfigure[Analysis of $\text{dim}~ z_*^i$ on MMM$2$]{
        \includegraphics[width=0.315\linewidth]{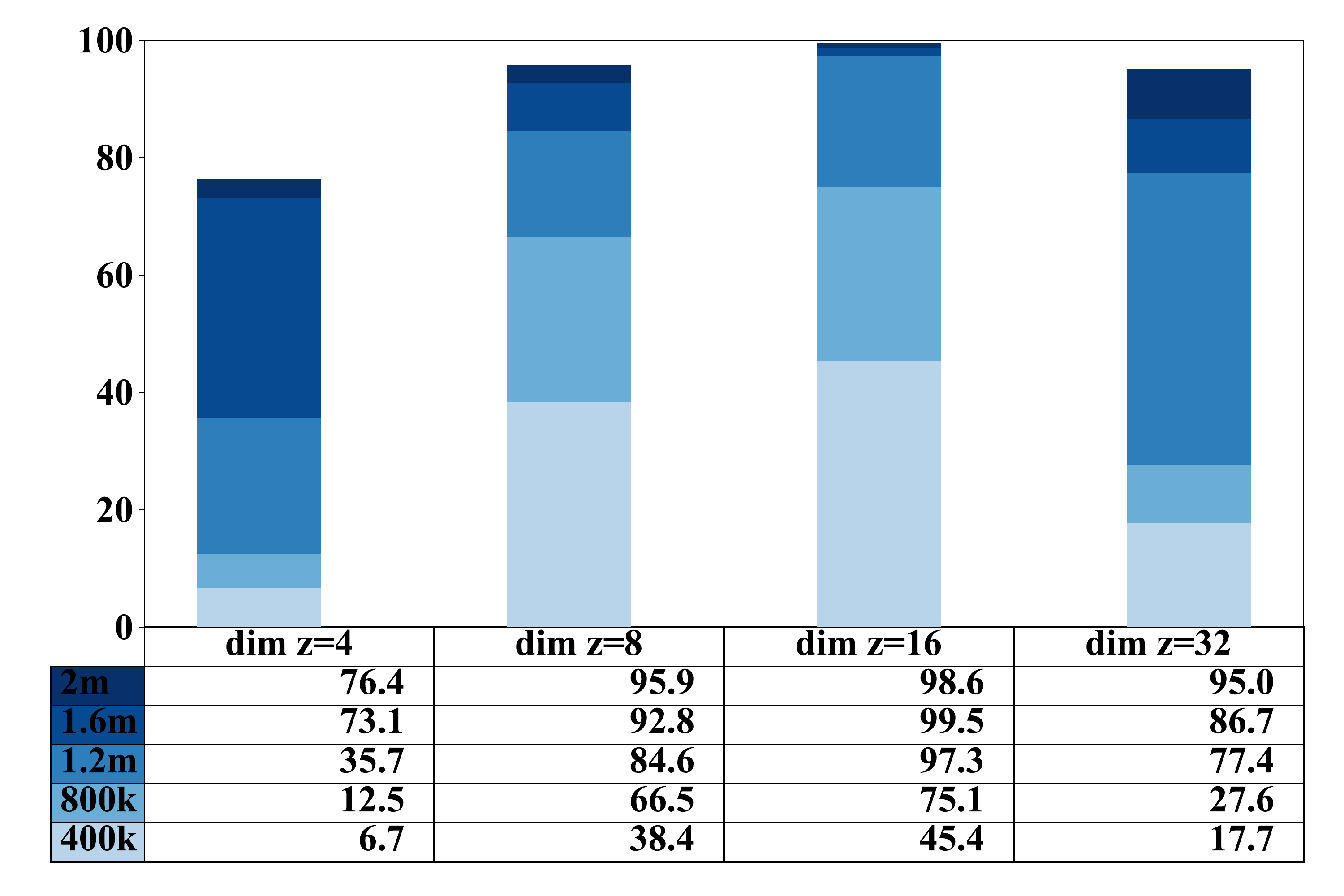}}
\caption{Analytical Evaluation.}\label{ablation}

\end{center}
\end{figure*}

To establish a better understanding of our method, we perform the following ablations: 1) HISMA-No-MI: by setting $\lambda_\text{MI}$ to zero (Equation (\ref{all-objs})); 2) HISMA-No-E: by setting $\lambda_\text{e}$ to zero and using $q_\phi$ instead of $F_\eta$ to sample next steps; 3) HISMA-No-R: ablates the term $J_m$ from the latent policy objective; 4) HISMA-No-$Q_R$: where we ablated the shared network $Q_R$ and broadcast $z_R$ to each agent's utility.  We also test the effect of both the segment length $s$ and the dimensionality of $z_*^i$ on the performance and report the results in Figure (\ref{ablation}). In addition to that, we show the effectiveness of our modified GAT module in learning complex interactive behaviors in Section (\ref{sec-viz}). 

Observing the analytical experiment on 6h\_vs\_8z (Figure \ref{ablation}), we found that our method performs optimally when  $s=50$. We believe that such value of $s$ enables consideration of delayed effects
of previous actions and capturing longer relations between observations than single-step models, it also achieves sufficient accurate predictions compared to models with larger number of steps. As for the dimensionality of a single latent strategy, we observe that the performance is sufficiently robust w.r.t the choice of $\text{dim}~z$ with a slight advantage to the value choice, $\text{dim}~z=16$.

\subsection{Overview on Learned Strategies}\label{sec-viz}
In Google Research Football (GRF), HISMA manages to uncover rewarding strategies that resemble real-life tactics in many cases. For example in the full game, we perceived two types of strategies: agent-related and team-related. The former manifests in the actions of one agent, such as taking a slight curve before shooting or sliding to tackle an opponent. As for team-related strategies, they represent interactions between more than one agent to achieve a task such as defensive pressure (closely marking a ball carrier to harass him into losing the ball). Examples of collective team behaviors include moving forward during a counterattack or switching to a defensive formation through other team attacks. One rewarding strategy HISMA found was a player making a forward run on the sideline, followed by a center pass creating a scoring opportunity. This is a well-known attacking tactic in real football matches, and it resulted in multiple goals and near-goal chances for our team.

In SC II environments, our latent policy learns useful strategies such that every agent’s strategy depends on: its past trajectories (health points, shield points, positions and actions of all agents, etc.), and its predicted future segments. The strategies automatically evolve during the progress of the game, adapting to changes and promoting better performances. 
For example, in 3s5z\_vs\_3s6z: the game starts with a plain team-related strategy: two Zealots disadvantageously attack the enemy’s 3 Stalkers, distracting them while the rest of the team battles the Zealots. When the distracting agents’ health drops, their relational strategy calls for backup, and other Zealots come for help. Influenced by their self-related strategy, damaged agents tend to retreat and stop engaging in battle. Supporting the latter behavior, a relational strategy emerges at the end of the game encouraging a Stalker to sacrifice itself battling 4 Zealots. We refer readers to Figures (\ref{strategy-viz}, \ref{strategy-viz1}) or the paper's accompanying videos for visualizations.

\subsection{Case Study: Trajectory Prediction Task}\label{sec-pred}

\begin{figure}[ht]
    \centering
    \subfigure[Truth \label{a}]{
        \fbox{\includegraphics[height=0.2\columnwidth,width=0.2\columnwidth]{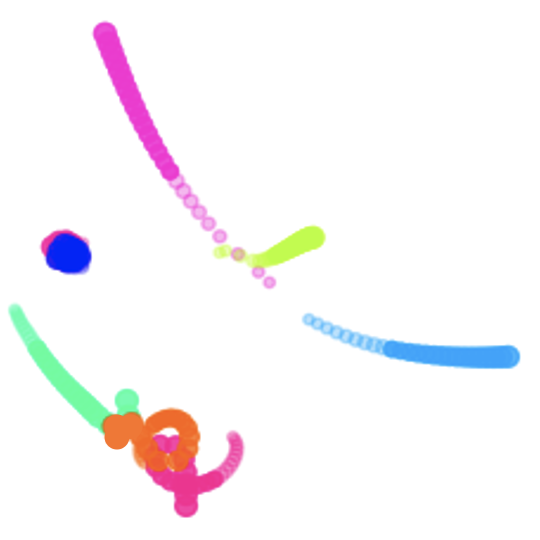}}}
       \subfigure[Full Model \label{b}]{
        \fbox{\includegraphics[height=0.2\columnwidth,width=0.2\columnwidth]{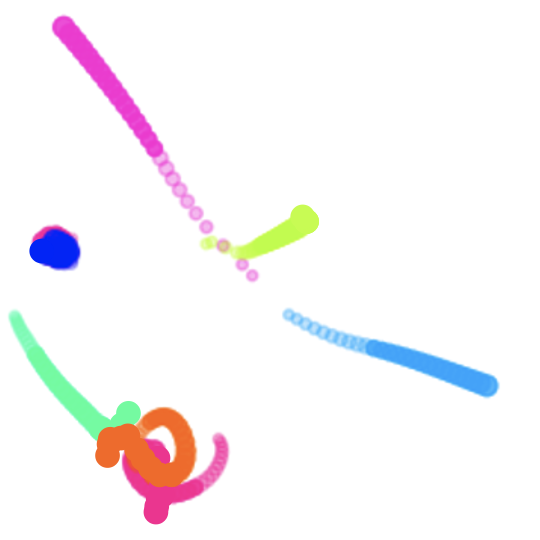}}}
        \subfigure[w/o $F_\eta$ \label{c}]{
        \fbox{\includegraphics[height=0.2\columnwidth,width=0.2\columnwidth]{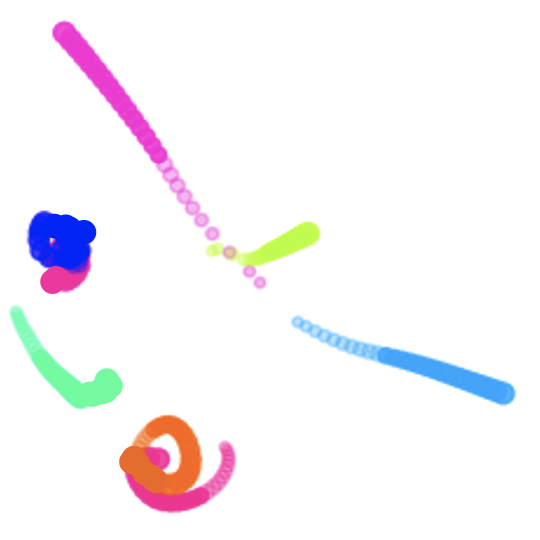}}}
        \subfigure[w/o GAT \label{d}]{
        \fbox{\includegraphics[height=0.2\columnwidth,width=0.2\columnwidth]{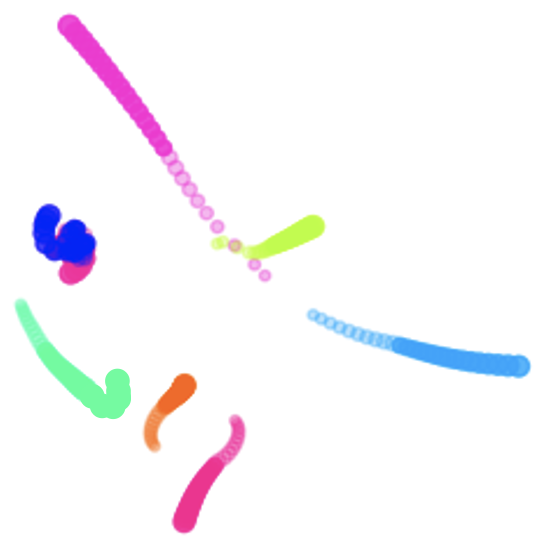}}}
\caption{Visualization of predicted trajectories on the Charges set.}
\label{pred}
\end{figure}

To test whether our trajectory forecasting model predicts the true trajectories and not some unrealistic representations of them, we experiment with the charged particles simulated system \citep{pmlr-v80-kipf18a},
which is controlled by simple physics rules. Details of the task and training along with the quantitative results and discussion can be found in Appendix \ref{traj-app}.

\section{Related Works}

There exists a rich literature of works on multi-agent reinforcement learning as it witnessed  significant advances in the recent years  \citep{Tan93multi-agentreinforcement,matignon2012independent,yang2020qatten,wen2018probabilistic,lowe2017multi,ryu2020multi, du2019liir,NEURIPS2020_8977ecbb,DBLP:journals/corr/abs-2006-04222}. Value-based algorithms decompose the joint value function into local utility functions in order to enable efficient optimization and decentralized execution, whereas VDN \citep{sunehag2018value}, QMIX \citep{rashid2018qmix}, QTRAN \citep{son2019qtran}, weighted-QMIX \citep{rashid2020weighted} have progressively enlarged the family of functions that can be represented by the mixing network. Most recently, \citet{wang2021qplex} introduced QPLEX, a dueling structure for representing  joint and individual  action-value functions. QPLEX significantly outperformed the aforementioned methods thus we use it in our experiments as a baseline.

\textit{Hierarchical MARL}. 
HISMA adopts a hierarchical structure where a high-level policy acts on a continuous latent space of strategies, and produces latent codes that can be later fed into agents' policies in order to define their behavior and help collect more rewards.  Similarly, several recent works addressed different challenges in MARL using specific hierarchies and achieved notable advancements: MAVEN \citep{mahajan2019maven} introduces a hierarchical control method with a shared latent variable encouraging committed, temporally extended exploration. 
ROMA \citep{wang2020roma} builds on QMIX by introducing shared \textit{role} encoder and decoder networks to encourage sub-task specialization and while ROMA accelerates training, \citet{chenghao2021celebrating} argue that parameter sharing cons become apparent in more complex tasks as agents tend to follow similar behaviors, which raises the need  for substantial exploration and foolproof strategies among agents. HSD \citep{yang2020hierarchical} provides each agent with a high-level policy to learn a skill, while an extrinsic team reward is used to conduct centralized training of high-level policies for cooperation.
Also, \citet{wang2021rode}  proposed RODE, a hierarchical method to decompose a complex task into a set of subtasks, each has a much smaller observation-action space and is associated with a role. A recent work that introduces a strategy-based based method in MARL COPA \citep{pmlr-v139-liu21m}, employs a hierarchical structure to learn strategies with maximum mutual information objective.

Although there has been extensive works on learning the model of the environment or predicting the future of a learning agent in single-agent reinforcement learning, the MARL literature extremely lacks works on this scope. To our best knowledge, HISMA is the first algorithm to incorporate the idea of predicting the future to produce self-consistent and high-rewarding strategies. With that being said, many algorithms have been proposed to investigate trajectory forecasting in multi-agent interacting systems in the framework of deep generative  models without optimization for rewards:  GRIN \citep{li2021grin} is CVAE that decodes the future from two latent codes: agents' intentions and social relations; EvolveGraph \citep{li2020evolvegraph} explicitly models interactions via dynamically evolving latent interaction graphs among agents; and many others \citep{graber2020dynamic,pmlr-v80-kipf18a,ma2021continual,zhan2018generating,kamra2020multi}. We conducted a case study on a simulated system and compared against the mentioned forecasting models  (See Section \ref{sec-pred}).

Since our work introduces multiple novel ideas in different scopes and due to page limit, we defer the discussion of more related works to Appendix \ref{more-related}.

\section{Closing Remarks}
In this paper, we introduced HISMA a novel MARL algorithm that makes use of a specific hierarchical structure and learning objectives to devise high-rewarding strategies that help predict the long-term future and plan according to it. We demonstrated its state-of-the-art performance by comparing it to powerful MARL approaches as HISMA managed to solve all of the given scenarios on the challenging GRF and SC II environments. We briefly discuss the limitations, impacts, and future works in Appendix \ref{discussion}.

\newpage

\newpage



\bibliography{example_paper}
\bibliographystyle{icml2022}

\newpage
\appendix
\onecolumn

\begin{figure*}[ht]
\begin{center}
\subfigure[Pressure: represents a relational strategy]{
        \includegraphics[width=0.48\textwidth]{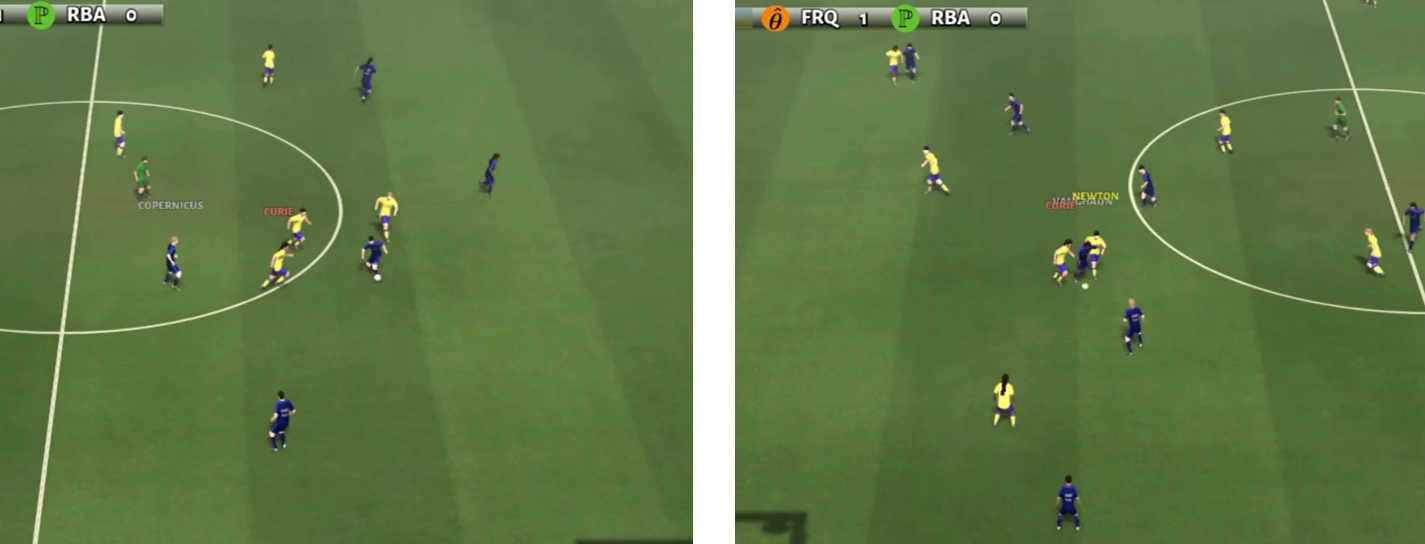}
    }
\subfigure[Sliding Tackle: represents an individual strategy]{
        \includegraphics[width=0.48\textwidth]{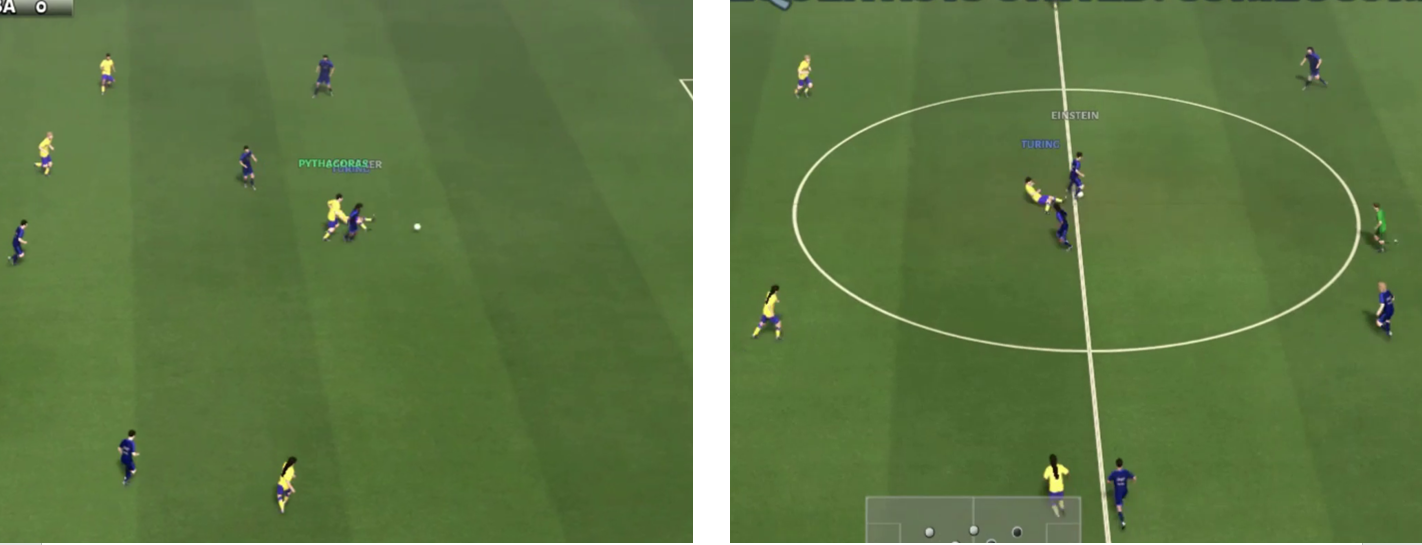}
    }

\subfigure[The yellow arrows are for players movement, blue arrows for ball movement, and red arrows for players movement on the map. Notice that the agent in the yellow square learned a strategy of camping to get a rebound when its teammate makes a shot attempt.]{
        \includegraphics[width=0.46\textwidth]{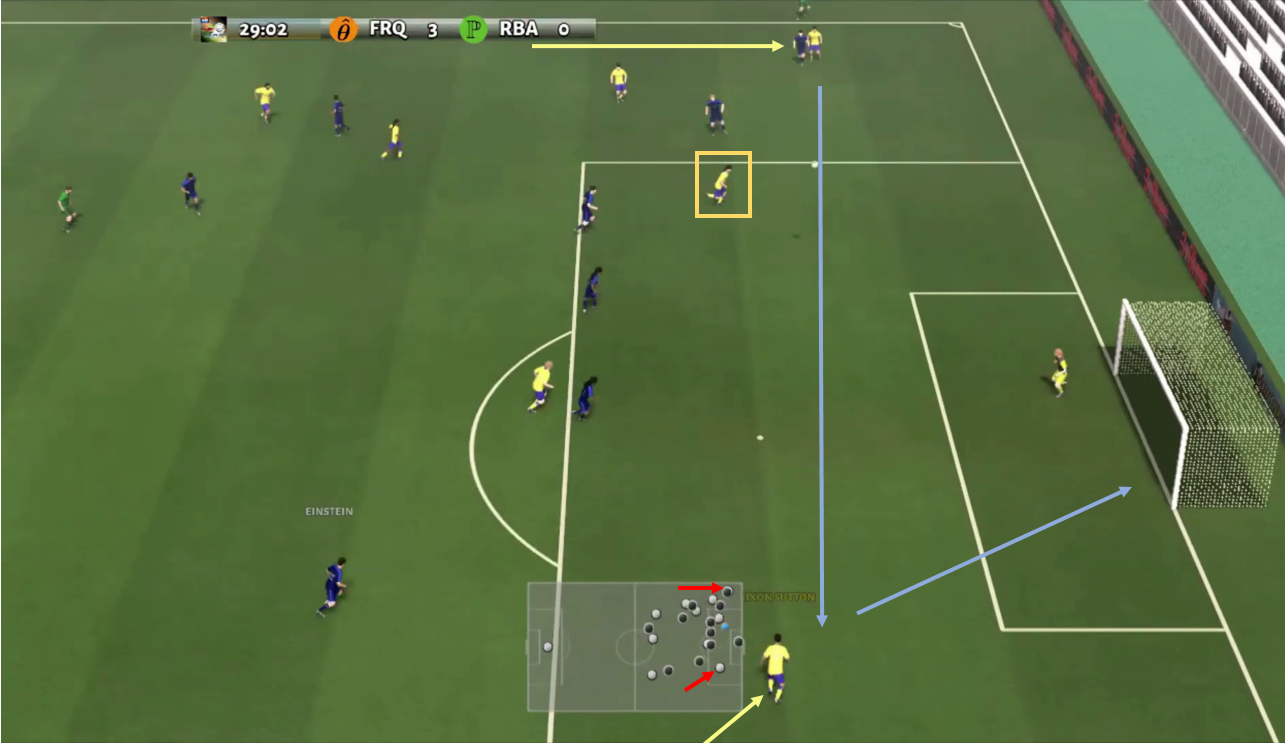}
    }
\subfigure[Counterattack]{
        \includegraphics[width=0.23\textwidth]{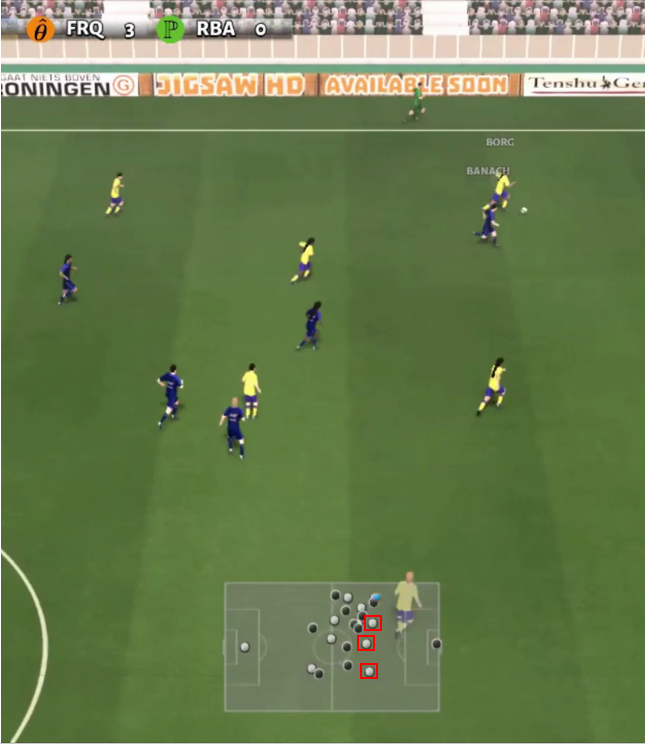}
    }
\subfigure[Defensive formation]{
        \includegraphics[width=0.23\textwidth]{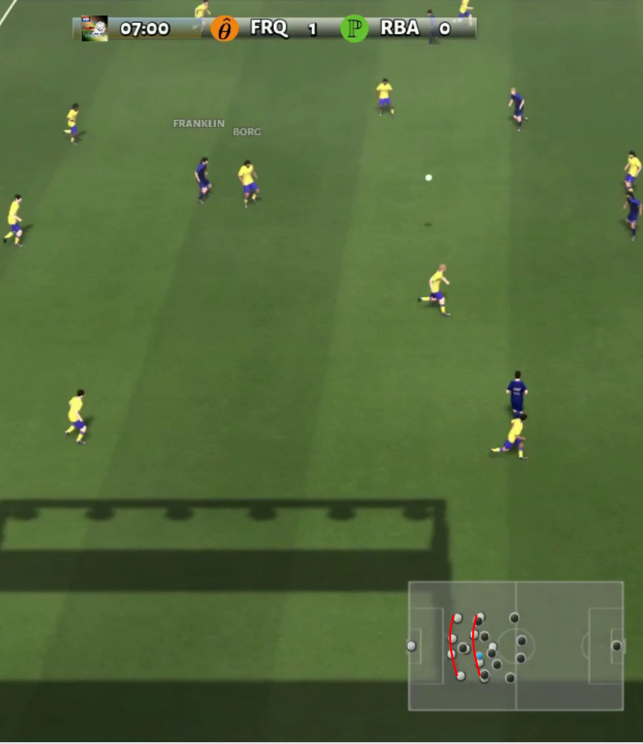}
    }
\caption{GRF strategies: notice on maps how players learn to spread to cover wide areas.}
\label{strategy-viz}
\end{center}
\vskip -0.2in
\end{figure*}

\begin{figure*}[ht]
\begin{center}
\subfigure[Two Zealots distracting enemy Stalkers]{
        \includegraphics[width=0.46\textwidth]{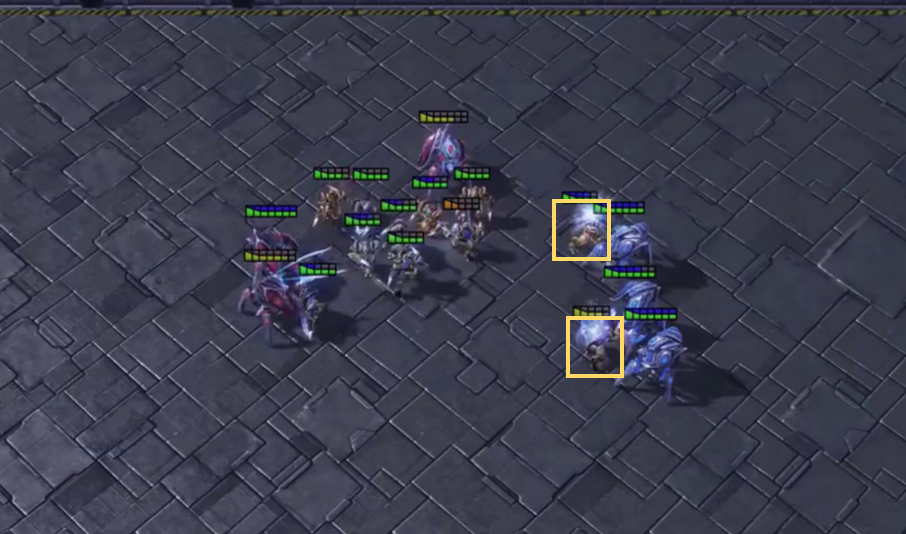}
    }
\subfigure[The Stalker in the top is battling enemy Zealots to give the bottom Stalkers time to heal.]{
        \includegraphics[width=0.46\textwidth]{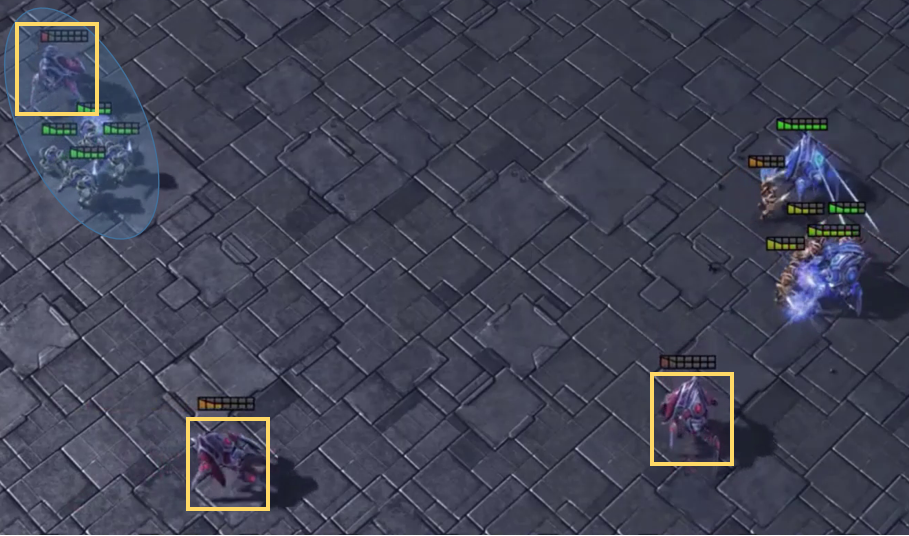}
    }
\caption{SMAC strategies}
\label{strategy-viz1}
\end{center}
\vskip -0.2in
\end{figure*}

\begin{figure}[t]
\begin{center}
\centerline{\includegraphics[width=1\linewidth]{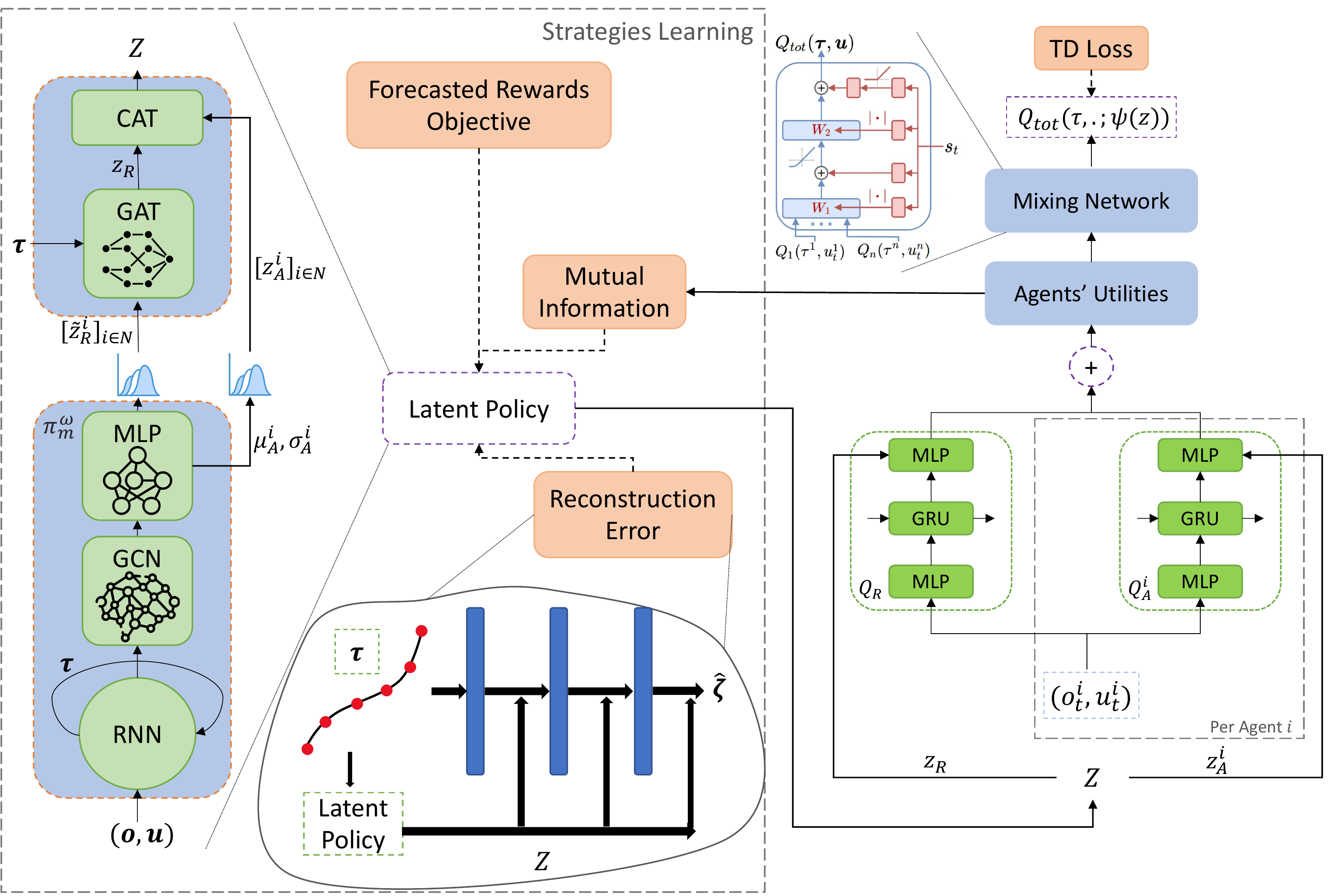}}
\caption{Full Illustration of HISMA}
\label{illus-big}
\end{center}
\vskip -0.2in
\end{figure}

\section{Mathematical Details}\label{math}
\begin{itemize}
    \item We show the derivation of lower bound of the MI objective by following the literature of variational inference:
\begin{align*}
    A&=I^{\pmb{\pi},\pi_m}(\pmb{\zeta}_k; z_k,\pmb{\tau}_{ks})
    =\mathbb{E}_{\pmb{\tau}_{ks}, z_k\sim\pi_m,\pmb{\zeta}_k\sim p(.|z_k,\pmb{\tau}_{ks})}
    \biggl[\log\frac{p\big(\pmb{\zeta}_k|z_k,\pmb{\tau}_{ks}\big)}{p(\pmb{\zeta}_k|\pmb{\tau}_{ks})}\biggl]\\
    &=\mathbb{E}\Biggl[\sum_{t=ks+1}^{(k+1)s} \log  \frac{p(\mathbf{a}_t|\pmb{\tau}_t;z_k)}{p(\mathbf{a}_t|\pmb{\tau}_t)}\cdot \frac{p(\mathbf{o}_{t+1}|\mathbf{a}_t,\pmb{\tau}_t;z_k)}{p(\mathbf{o}_{t+1}|\mathbf{a}_t,\pmb{\tau}_t)}\Biggl]\\
    &= \sum_{t=ks+1}^{(k+1)s} \Bigg(\mathbb{E}\biggl[\log  \frac{\sigma(\mathbf{a}_t|\pmb{\tau}_t;z_k)}{p(\mathbf{a}_t|\pmb{\tau}_t)} \cdot \frac{q_\phi(\mathbf{o}_{t+1}|\mathbf{a}_t,\pmb{\tau}_t;z_k)}{p(\mathbf{o}_{t+1}|\mathbf{a}_t,\pmb{\tau}_t)}\biggl] \\
    &~~~~~~~~~~~~~~~~~~~~~~~~~~~~~~~~~~~~~~~~~~~~~~~~~~~~+\text{KL}[p(\mathbf{a}_t|\pmb{\tau}_t;z_k)||\sigma(\mathbf{a}_t|\pmb{\tau}_t;z_k)]
    +\text{KL}[p(\mathbf{o}_{t+1}|\mathbf{a}_t,\pmb{\tau}_t;z_k)||q_\phi(\mathbf{o}_{t+1}|\mathbf{a}_t,\pmb{\tau}_t;z_k)]
    \Bigg)\\
    &\geq \mathbb{E}\Biggl[\sum_{t=ks+1}^{(k+1)s} \log  \frac{\sigma(\mathbf{a}_t|\pmb{\tau}_t;z_k)}{p(\mathbf{a}_t|\pmb{\tau}_t)} \cdot \frac{q_\phi(\mathbf{o}_{t+1}|\mathbf{a}_t,\pmb{\tau}_t;z_k)}{p(\mathbf{o}_{t+1}|\mathbf{a}_t,\pmb{\tau}_t)}\Biggl] \qquad\qquad\qquad\qquad\qquad \qquad\qquad\qquad\qquad(\text{KL}[\cdot||\cdot]\geq 0)
\end{align*}
This proves inequality (\ref{bound1}) which is similar to (\ref{bound2}). The Kullback–Leibler divergence terms translate to the objectives $D_\phi$ and $D_\xi$, whereas minimizing them tightens the bounds in order to yield more accurate estimates of $J_\text{MI}^{\pmb{\pi},k}$.

\end{itemize}

\begin{figure}[t]
\begin{center}
\subfigure[GRF Corner]{
        \includegraphics[width=0.315\textwidth]{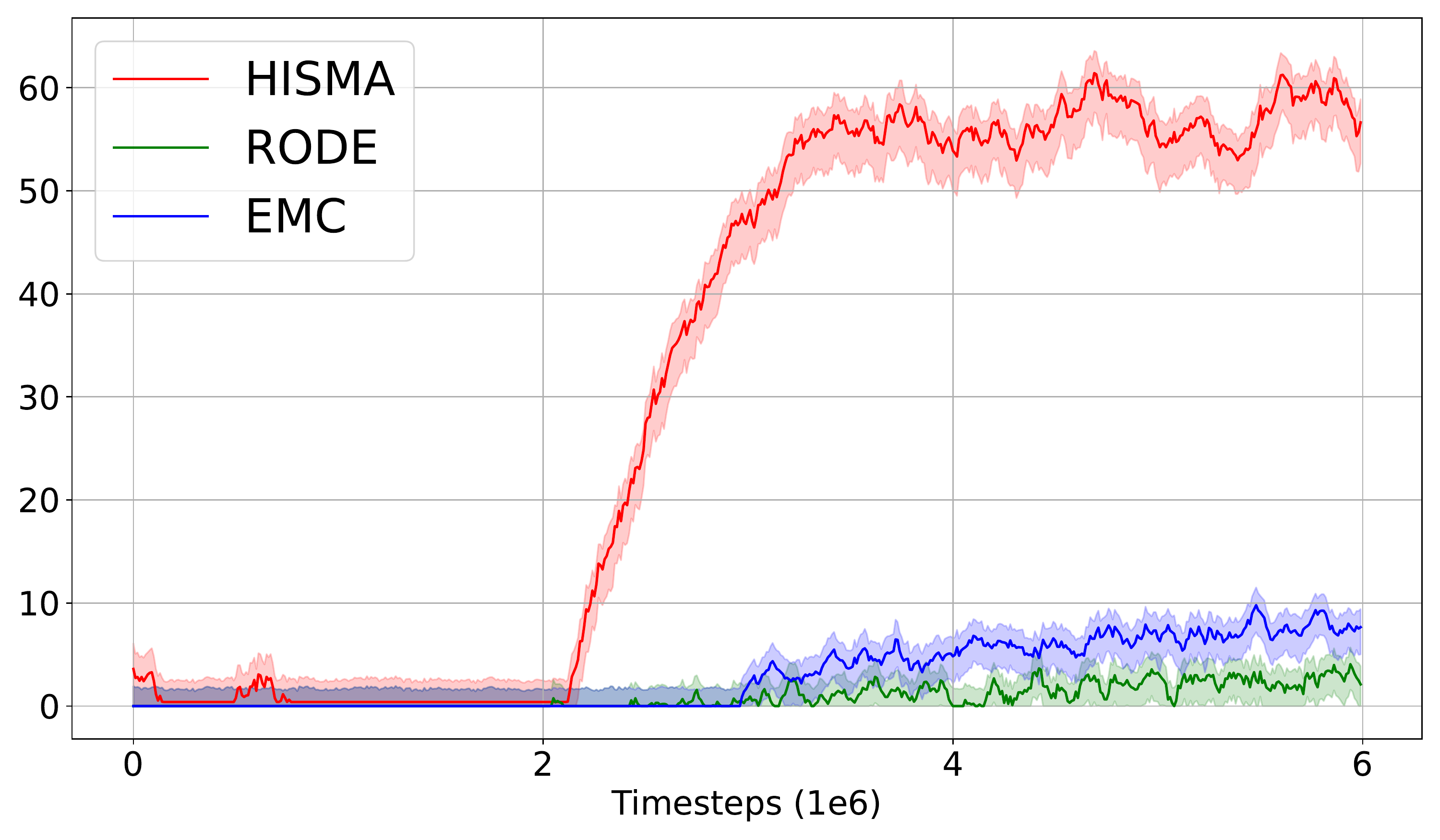}
    }
\subfigure[SC II 5s10z]{
        \includegraphics[width=0.315\textwidth]{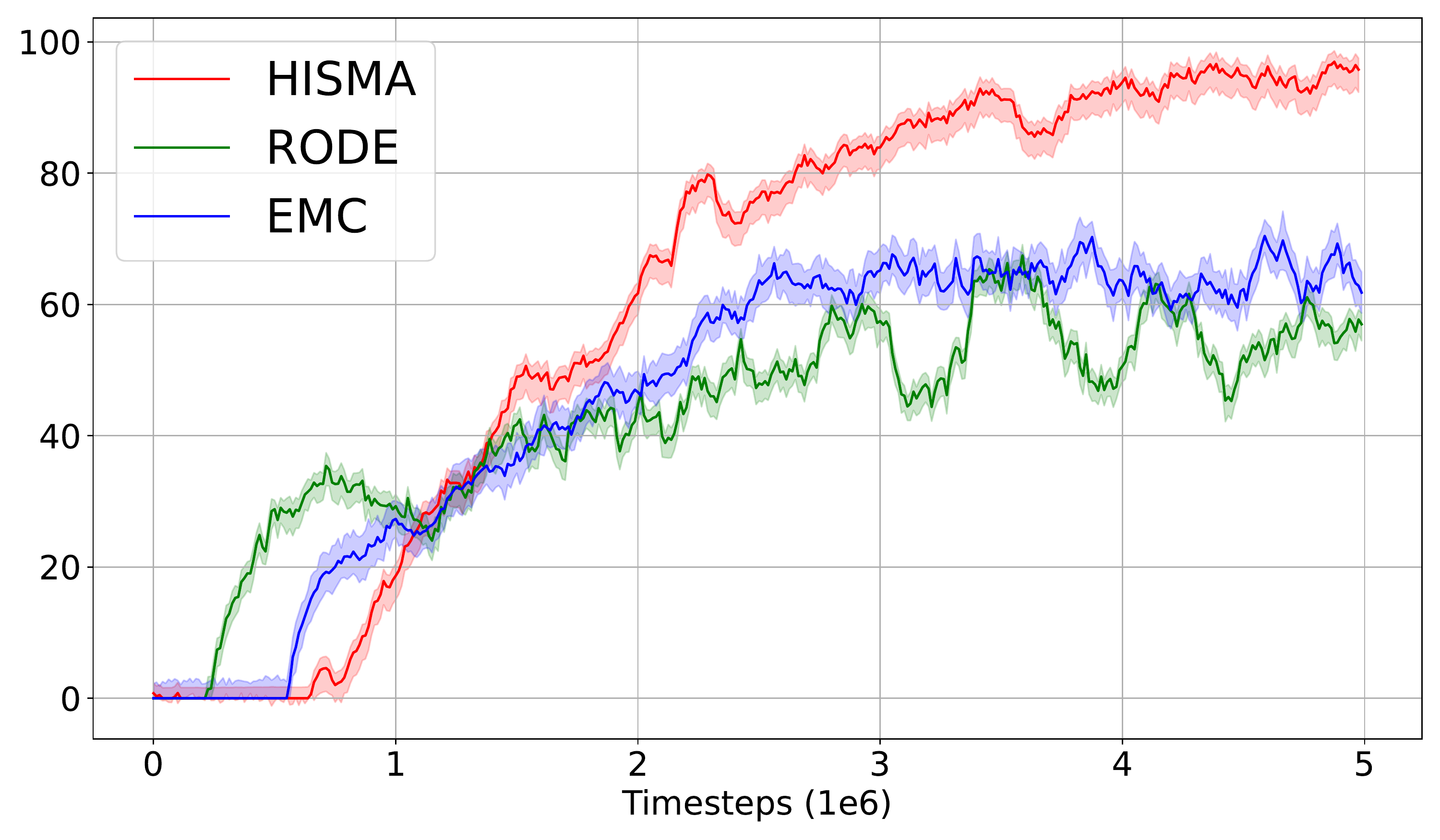}
    }
\subfigure[SC II 6h\_vs\_8z]{
        \includegraphics[width=0.315\textwidth]{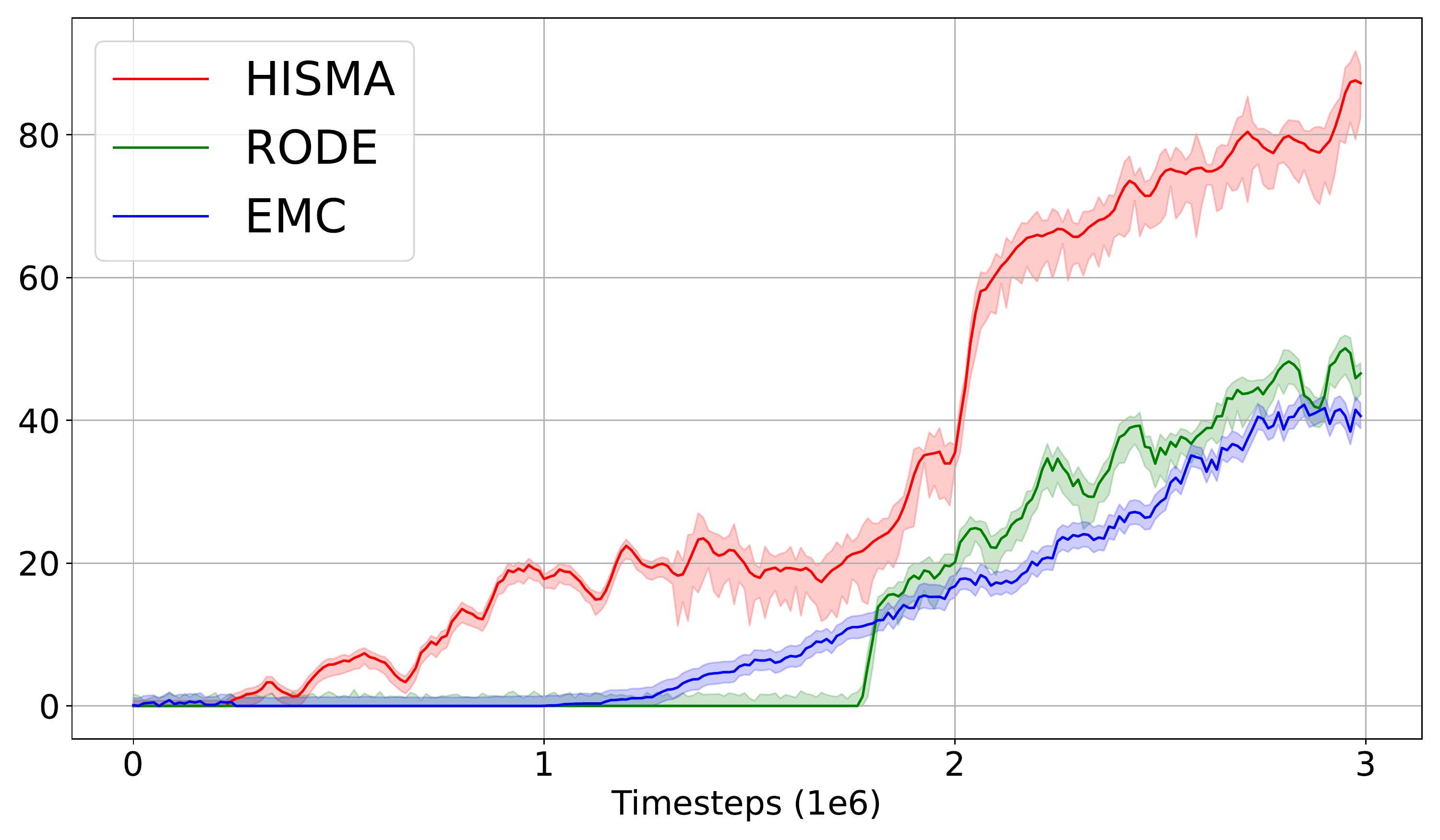}
    }
\subfigure[SC II 2c\_vs\_64zg \label{fuuuck}]{
        \includegraphics[width=0.315\textwidth]{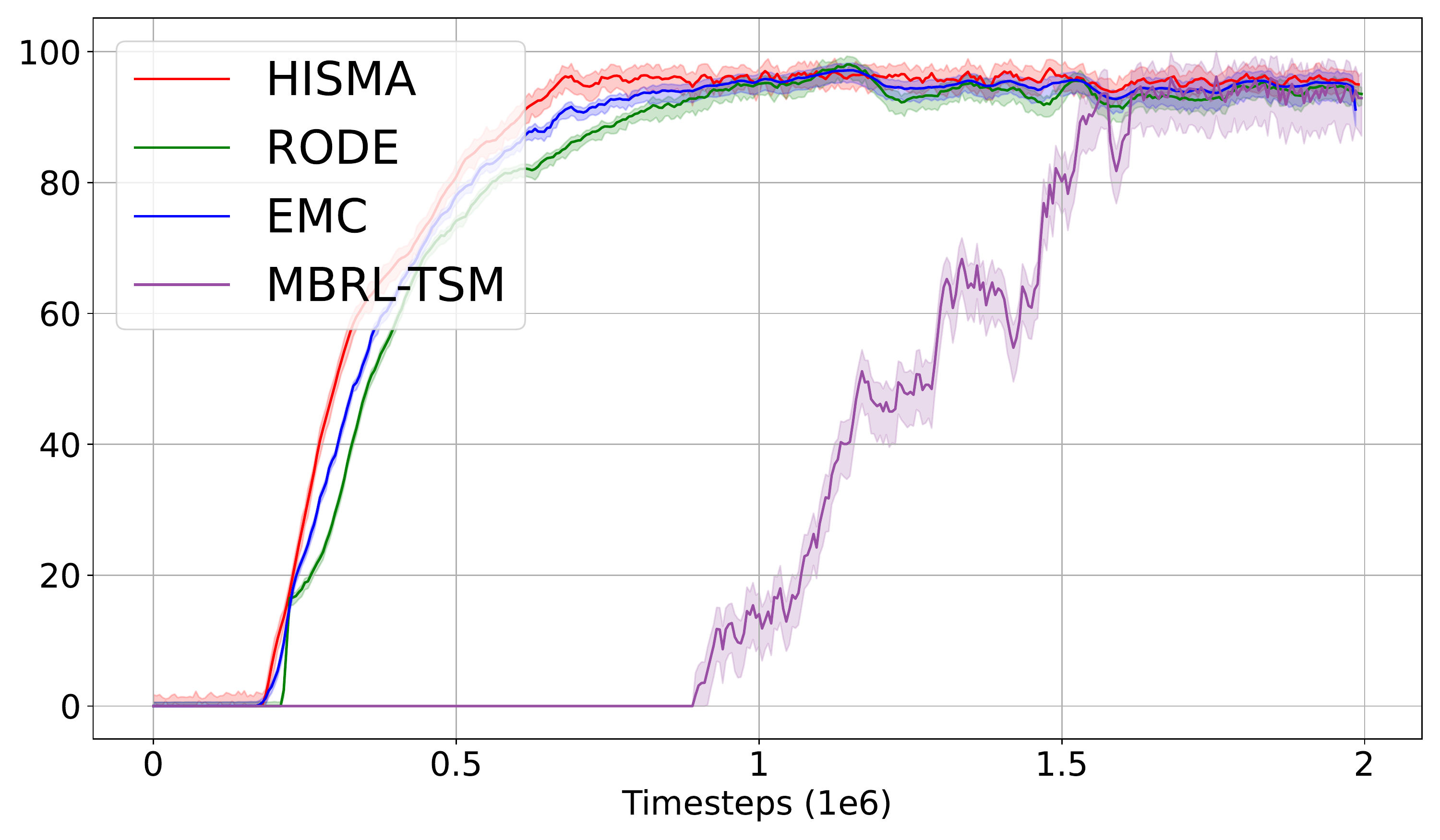}
    }
\subfigure[GRF run to score w/ keeper]{
        \includegraphics[width=0.315\textwidth]{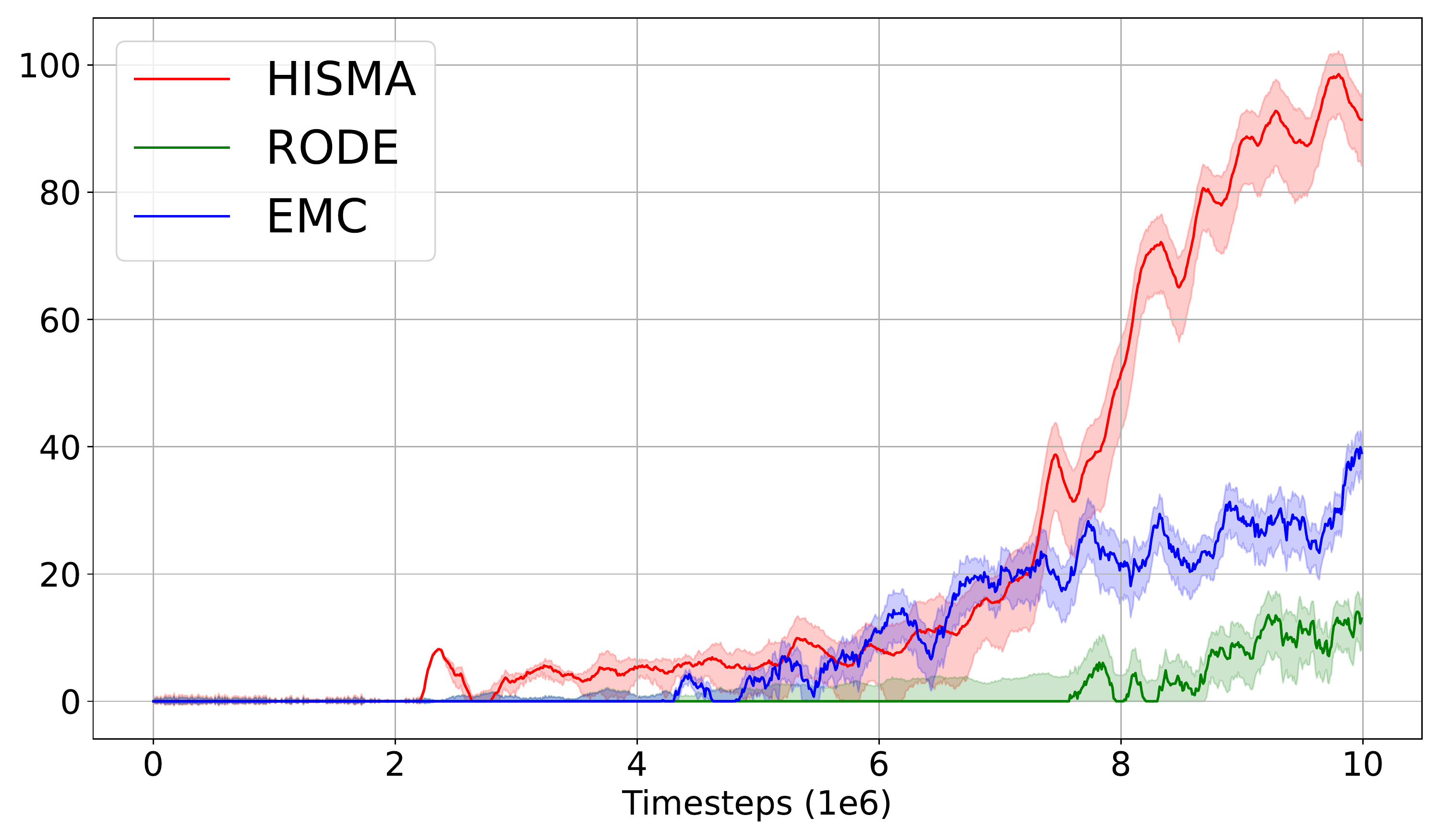}
    }
\caption{Additional Experiments on Additional Baselines (EMC, RODE, MBRL-TSM)}
\label{more-smac}
\end{center}
\end{figure}

%
\begin{algorithm}[ht]
   \caption{HISMA}
   \label{alg}
\begin{algorithmic}
   \STATE Initialize $s,\theta,\omega,\eta,\phi,\xi,\mathcal{D}=\{\}$
   \FOR{each episode:}
   \FOR{$t=1:T$}
   \FOR{each gradient step}
   \STATE $\{\pmb{\tau}_{k_ls},\pmb{\zeta}_{k_l}\}_1^{l}\sim\mathcal{D}$
   \STATE $\Delta \phi \leftarrow \lambda_\phi \hat{\nabla}_{\phi} J_{\text{MI}} - \lambda_\phi \hat{\nabla}_{\phi} D_\phi$
   \STATE $ \Delta \xi \leftarrow \lambda_\xi \hat{\nabla}_{\xi} J_{\text{MI}} - \lambda_\xi \hat{\nabla}_{\xi}D_\xi$
   \STATE $\Delta \theta \leftarrow \lambda_\theta \hat{\nabla}_\theta J_{\text{MI}} - \lambda_\theta\hat{\nabla}_\theta \mathcal{L}$
   
   \IF{$t~ \text{mod} ~s=0$}
   \STATE $ \Delta\eta \leftarrow -\lambda_\eta\hat{\nabla}_{\eta}J_e$
   \STATE $\Delta \omega \leftarrow \lambda_\omega \hat{\nabla}_\omega J_{\text{MI}}+\lambda_\omega\hat{\nabla}_\omega J_m-\lambda_\omega\hat{\nabla}_\omega J_e$
   
   \ENDIF
   \ENDFOR
   \STATE $\mathcal{D}\leftarrow(\pmb{o}_t,\pmb{u}_t,r_t,\pmb{o}_{t+1})$
   \ENDFOR
   
   \ENDFOR
\end{algorithmic}
\end{algorithm}

\section{Additional Experimental Results}\label{add-exps}

\subsection{Comparison to Multi-Agent Model-based Temporal Segment Models (MBRL-TSM)}

We compare HISMA against MBRL-TSM algorithm \citep{krupnik2020multi} on the 2c\_vs\_64zg SC II scenario and report results in Figure (\ref{fuuuck}). Note that MBRL-TSM is a 2-agent method and thus was not used as a baseline in other scenarios.

\subsection{Trajectory Prediction Task Details}\label{traj-app}
\begin{table}[h!]
\caption{Performance against Trajectory Prediction Baselines}
\label{trajectory-pred}
\begin{center}
\begin{small}
\begin{sc}
\begin{tabular}{lcc}
\toprule
Model & ADE & FDE \\
\midrule
Ours (Full Model)  & \textbf{0.288} &\textbf{0.955} \\
Ours (w/o Prediction Model $F$) &0.315 & 1.138\\
Ours (w/o GAT) & {0.467} &1.663 \\
GRIN &0.394 &1.209 \\
EvolveGraph (Double Stage) &0.407 &1.221\\
FQA &0.420 & 1.218\\
NRI (Dynamic) &0.565 & 1.766\\
\bottomrule
\end{tabular}
\end{sc}
\end{small}
\end{center}
\vskip -0.1in
\end{table}
We assume a given scene with the spatial coordinates $\mathbf{p}_t^
i = (x_t^
i, y_t^
i )$ of all agents $i\in \mathcal{N}$.
The task is to observe the agents for $T_{obs}$ time steps and infer their behavior for the rest of the time horizon (i.e. $T_{obs}+1$ to $T$). 
We generate 15K scenes
for training, and 3K each for validation and test respectively. For each scene, we set the horizon $T=25$ and the number of observed time-steps $T_{obs}=10$. For training, We use the latent policy to generate the agents' latent behaviors $z=[z_A,z_R]$ by maximizing the MI objective: $\sum_k\gamma^{ks}J_\text{MI}^k$, we then use $F_\eta$ to decode the future from $z$ and $\pmb{\tau}_{T_{obs}}$ by minimizing the reconstruction loss. For evaluation, we use the average minimum displacement error (ADE) and the final minimum displacement error (FDE)  metrics and  compare HISMA's performance to the cutting-edge methods in multi-agent trajectory prediction: GRIN \citep{li2021grin}, EvolveGraph \citep{li2020evolvegraph}, FQA \citep{kamra2020multi}, dNRI \citep{graber2020dynamic}. It is important to note that these baselines were not included in the SC II and GRF experiments since they only aim to solve the forecasting problem and do not have a specific architecture for reward sum maximization to incorporate reward-seeking learning agents. 


A prominent advantage of our method that leads to more accurate predictions is that along learning to extract the interaction features between the particles thanks to our modified GAT module, it learns to generate latent (variables/representations) that minimize the uncertainty of the future and predict it by learning to filter out unnecessary information about agent's self-behaviors and relationships.

In Figure (\ref{pred}), we visualize the ground truth trajectories (\ref{a}), predicted trajectories of our full model (\ref{b}), predicted trajectories of our model after ablating the model $F_\eta$ and using $q_\phi$ to sample next steps (\ref{c}), and predicted trajectories of our full model without our modified GAT module (\ref{d}). Note that (b) and (c) output accurate predictions with a slight advantage to our full model. However, (d) demonstrates its inability to model the complex swirling (pink and orange charges) due to the unavailability of sufficient information about interaction features necessary to model such complex behaviors. 


\section{Implementation Details}\label{hyper}
All experiments were performed on  a high performance computing system with a SLURM \citep{yoo2003slurm} job scheduler. The compute nodes used each has two NVIDIA Volta V100 GPUs, and a dual socket Intel Xeon Gold 6248 processor with 20 cores each socket.

In this paper, we base our algorithm on QMIX \citep{rashid2018qmix}. Each agent has a neural network to approximate its local utility. The local utility network consists of three layers, a fully-connected layer, followed by a 64 bit GRU, and followed by another fully-connected layer that also takes the individual strategy as an input and outputs
an estimated local value for each action. This output is added to the output of a similar shared network, but with one difference: that is the shared network takes all the relational strategies as input to its second MLP. The  utilities are fed into a mixing network estimating the global action value. The mixing network has a 32-dimensional hidden layer with ReLU activation. Parameters of the mixing network are generated by a hyper-net conditioning on the global state. This hyper-net has a fully-connected hidden layer of 32 dimensions. These settings are the same as QMIX.

The optimization is conducted using
RMSprop with a learning rate of $5 \times 10^{-4}$, $\alpha$ of $0.99$, and with no momentum or weight decay.  We run 8 parallel environments to collect samples. Batches of $32$ episodes are sampled from the replay buffer, and the whole framework is trained end-to-end on fully unrolled episodes. Neural networks were implemented using PyTorch framework v1.8.1+cu102, and
graph neural networks include graph attention neural networks and graph convolution networks are
defined using DGL \citep{wang2019deep} v0.6.1.

\section{Full report on Related Works}\label{more-related}

To alleviate the partial observability problem, \citep{cao2021linda} proposed LINDA---a method of building awareness for agents in order to efficiently  leverage local information. 
SVO \citep{mckee2020social}, on the other hand, is an algorithm that introduces diversity into heterogeneous agents for more generalized and high-performing policies in social dilemmas. CMAE \citep{liu2021cooperative} promotes cooperative exploration by equipping the agents with a common goal and training them to coordinate behaviors to reach it, which corresponds to our data gathering process where agents celebrate surprising encounters and learn to collect diverse data to learn the true dynamics and accurately predict the future and plan accordingly.
Recently, policy gradient MARL algorithms: FOP \cite{ pmlr-v139-zhang21m}, DOP \citep{wang2020dop}, MADDPG \citep{lowe2017multi}, FACMAC \citep{peng2021facmac}, HAMA \citep{ryu2020multi}, and others \citep{zhang2022scc,su2021value,DBLP:journals/corr/abs-2003-06709,iqbal2019actor} investigated the decomposed actor-critic framework. In particular, FOP advances DOP by proposing to factorize the optimal joint policy obtained from maximum entropy MARL into individual
policies with theoretical guarantees of convergence to the global optimum. FACMAC combines MADDPG with QMIX but relaxes the monotonic constraints on the factorized critics. HAMA employs a hierarchical GAT to learn state representation for later use in AC optimization in small-scale multi-agent environments.

In single-agent RL, many works studied the idea of incorporating the future or learning the dynamics to make more informed decisions \citep{buckman2018sample,sutton2018reinforcement,holland2018effect,tang2019multiple,racaniere2017imagination,peters2010relative,daniel2012hierarchical,ding2020mutual}: \citet{mishra2017prediction} proposed two VAEs  to predict the future sequence of states and perform a segment-based policy optimization over the latent space. However, their method has limited scalability due to the unavailability of an exploration mechanism. Later, \citet{krupnik2020multi} extended the former work to two-agent competitive and cooperative settings and used a disentangled generative model to learn the dynamics. \citet{nachum2018nearoptimal} connects connects mutual information estimators to representation learning in hierarchical RL. Following on that,
The work of \citet{kim2019emi} aims to learn representations of states and actions so that transition dynamics are linear, and the agent is rewarded for stumbling upon "surprising" transitions that are not captured by the model in order to improve exploration. 
\citet{ke2018modeling} focused on predicting a future observation-action sequence by modeling their distribution using latent codes and a regularized form of ELBO \citep{goyal2017z}. Consequently, they use model predictive control (MPC) \citep{mayne2000constrained} to perform planning over the latent space. On the other hand, \citet{co2018self} introduced SeCTAR, a bottom-up hierarchical RL method to learn representations of an agent's trajectories and reason about the future outcome of the model.  \citet{eysenbach2018diversity,Lee2020Learning,Sharma2020Dynamics-Aware,gehring2021hierarchical} investigate the framework of discovering diverse skills (without extrinsic reward in some cases) by maximizing
entropy as well as mutual information between resulting states and latent representations of skills.
 We precisely discuss COPA \citep{pmlr-v139-liu21m}, it aims at solving dynamic team composition problems, creating many fundamental differences between us: first, COPA assumes an agent with an access to the global state information (the coach). Second, we propose a strategy that is learned using GAT \citep{velivckovic2018graph}, while COPA applies the multi-head attention mechanism for each agent. Third, apart from the mutual information objective, we introduce a trajectory prediction architecture that improves strategy learning. We show our empirical advantage over COPA in section (\ref{experiments}).One remotely related work is EOI \citep{jiang2021emergence}, where agents are intrinsically rewarded for being correctly predicted by a probabilistic classifier that is learned based on agents’ observations.
More related, CDS \citep{chenghao2021celebrating} maximizes the mutual information between individual trajectories and agents’ identities to encourage the specialty of individual trajectories. We share with CDS the idea of equipping each agent with two Q-functions, a local and a shared Q-function, with the exception that we condition local Qs on individual strategies, and shared Q on shared strategies.

In addition, many other works study the multi-agent problem from the perspective of  communication \citep{kim2021communication,Wang*2020Learning,singh2018individualized,foerster2016learning}, influence \citep{Wang*2020Influence-Based,jaques2019social,oliehoek2012influence},  coordination graphs \citep{bohmer2020deep,pmlr-v80-bargiacchi18a,grover2018evaluating,yang2018glomo,pmlr-v80-kipf18a}, etc.

\section{SC II Description}\label{smac-app}
The StarCraft II unit micromanagement task is considered as one of the most challenging cooperative multi-agent testbeds
for its high degree of control complexity and environmental stochasticity.  SC II (or SMAC) provides a rich set of units each with diverse actions, allowing for extremely complex cooperative behaviors among agents. We thus evaluate our method on several SC II micromanagement tasks from the SMAC benchmark  \citep{samvelyan2019starcraft}, where a group of mixed-typed units controlled by decentralized agents needs to cooperate to defeat another group of mixed-typed enemy units controlled by built-in heuristic rules with “difficult” setting; the battles can be both symmetric (same units in both groups) or asymmetric. Each agent observes its own status and, within its field of view, it also observes other units’ statistics such as health, location, and unit type (partial observability); agents can only attack enemies within their shooting range. A shared reward is received on battle victory as well as damaging or killing enemy units. Each battle has step limits set by SMAC and may end early.

\begin{table}
	\centering
\scalebox{1}{\begin{tabular}{ccc}
	\toprule 
	Map Name & Ally Units & Enemy Units \\ \hline
	2s3z &  2 Stalkers \& 3 Zealots &  2 Stalkers \& 3 Zealots  \\ 
	3s5z &  3 Stalkers \& 5 Zealots &  3 Stalkers \& 5 Zealots \\ 
	1c3s5z & 1 Colossus, 3 Stalkers \& 5 Zealots &  1 Colossus, 3 Stalkers \& 5 Zealots \\ \hline
	5m\_vs\_6m & 5 Marines & 6 Marines \\
	10m\_vs\_11m & 10 Marines & 11 Marines \\
	27m\_vs\_30m & 27 Marines & 30 Marines \\
	3s5z\_vs\_3s6z & 3 Stalkers \& 5 Zealots &  3 Stalkers \& 6 Zealots \\
	MMM2 & 1 Medivac, 2 Marauders \& 7 Marines & 1 Medivac, 2 Marauders \& 8 Marines \\ \hline
	2s\_vs\_1sc & 2 Stalkers & 1 Spine Crawler \\
	3s\_vs\_5z & 3 Stalkers & 5 Zealots \\
	6h\_vs\_8z & 6 Hydralisks & 8 Zealots \\
	bane\_vs\_bane & 20 Zerglings \& 4 Banelings & 20 Zerglings \& 4 Banelings \\
	2c\_vs\_64zg & 2 Colossi & 64 Zerglings \\ 
	corridor & 6 Zealots & 24 Zerglings \\ \hline
	5s10z & 5 Stalkers \& 10 Zealots &  5 Stalkers \& 10 Zealots \\ 
	7sz & 7 Stalkers \& 7 Zealots &  7 Stalkers \& 7 Zealots \\ 
	1c3s8z\_vs\_1c3s9z & 1 Colossus, 3 Stalkers \& 8 Zealots &  1 Colossus, 3 Stalkers \& 9 Zealots \\ \toprule
\end{tabular}}
\caption{The complete SMAC benchmark \citep{samvelyan2019starcraft} (adapted from \citet{wang2021qplex}'s paper).}\label{smacenvs}\end{table}

\section{Discussion}\label{discussion}

\textbf{Potential Limitations:} Although our comprehensive experiments show unprecedented results on challenging scenarios, we noticed that HISMA is sensitive to the choice of the number of prediction steps $s$.
We found an optimal value of $s=50$, as lesser values lead to momentary strategies that trivialize the essence of future prediction. For bigger $s$ values, our model fails to yield unbiased estimates of future segments because of the complexity of the environments. On the other hand, although HISMA surpasses all baselines on the scenario 27m\_vs\_30m, one can notice the slow learning at the beginning. One possible explanation is that the extra complexity (including the consistent exploration) of the method can harm performance in some cases.

\textbf{Scientific \& Societal Impacts:} This work shed a light on an effective class of algorithms in MARL with many novel ideas and structures which, we believe, might be of great interest to the MARL research community. Particularly, defining a strategy through a crafted information-theoretic measure as well as learning dynamic relations between acting agents, not to mention the well-established data gathering process, proved to be effective and powerful. Multi-agent deep reinforcement learning has potential applications in various areas such as autonomous
vehicles, robotic warehouses, smart grids, and more. Our research could be used
to improve multi-agent reinforcement learning in such applications. However, it must be noted that
real-world application of MARL algorithms is currently not viable due to open problems in AI
explainability, robustness to failure cases, legal and ethical aspects, and other issues, which are
outside the scope of our work. In addition, this work does not have any foreseeable negative societal impacts at the moment.  

\textbf{Future Works} might investigate adding behavior policies for offline data collection in order to enhance exploration.
Another possibility is to enable learning strategies in continuous action spaces by combining HISMA with an actor-critic method such as FOP.

\end{document}